\documentclass[journal]{IEEEtran}
\usepackage[utf8]{inputenc}
\usepackage{float}
\usepackage{graphicx}
\usepackage{subcaption}
\usepackage{textcomp}
\usepackage{xcolor}
\usepackage{float}
\usepackage{multirow}
\usepackage{csquotes}
\usepackage{url}
\usepackage{amsmath}

\usepackage{tabularx}
\usepackage{booktabs}
\newcolumntype{Y}{>{\raggedleft\let\newline\\\arraybackslash\hspace{0pt}}X}
\newcolumntype{Z}{>{\centering\let\newline\\\arraybackslash\hspace{0pt}}X}

\usepackage{enumitem}
\setlist[itemize]{noitemsep,topsep=0pt}

\ifCLASSINFOpdf

\else

\fi

\begin{document}

\title{Single Morphing Attack Detection using Feature Selection and Visualisation based on Mutual Information}

\author{Juan Tapia,~\IEEEmembership{Member,~IEEE,}
        and~Christoph Busch,~\IEEEmembership{Member,~IEEE,}\\ 
   **The following paper is a pre-print. The publication is currently under review for IEEE.**     
\thanks{Juan Tapia and Christoph Busch, da/sec-Biometrics and Internet Security Research Group, Hochschule Darmstadt, Germany, e-mail: (juan.tapia-farias@h-da.de, christoph.busch@h-da.de).}

\thanks{Manuscript received xxx; revised xx.}}

\markboth{Journal of \LaTeX\ Class Files,~Vol.~14, No.~8, August~2015}%
{Shell \MakeLowercase{\textit{et al.}}: Bare Demo of IEEEtran.cls for IEEE Journals}

\maketitle

\begin{abstract}
Face morphing attack detection is a challenging task. Automatic classification methods and manual inspection are realised in automatic border control gates to detect morphing attacks. Understanding how a machine learning system can detect morphed faces and the most relevant facial areas is crucial. Those relevant areas contain texture signals that allow us to separate the bona fide and the morph images. Also, it helps in the manual examination to detect a passport generated with morphed images. This paper explores features extracted from intensity, shape, texture, and proposes a feature selection stage based on the Mutual Information filter to select the most relevant and less redundant features. This selection allows us to reduce the workload and know the exact localisation of such areas to understand the morphing impact and create a robust classifier. The best results were obtained for the method based on Conditional Mutual Information and Shape features using only 500 features for FERET images and 800 features for FRGCv2 images from 1,048 features available. The eyes and nose are identified as the most critical areas to be analysed. 
\end{abstract}



\maketitle

\section{Introduction}
\IEEEPARstart In recent years, ID verification systems have been exposed to variations of presentation attacks. For instance, they compare the user selfie with a picture of the photo ID extracted from the user ID card or passport, where the critical challenge becomes ensuring whether or not the ID card image has been tampered with in the digital or physical domain. Image tampering is a significant issue for such scenarios and biometric systems at large \cite{Ferrara}.

One of these approaches is related to the passports, and the Morphing attack on face recognition systems based on the enrolment of a morphed face image, which is averaged from two-parent images and allowing both contributing subjects to travel with the passport \cite{Ferrara, Scherhag_survey, ScherhagDeep}. 
Morphing attack detection is a new topic aimed to detect unauthorised individuals who want to gain access to a "valid" identity in other countries. Morphing can be understood as a technique to combine two o more look-alike facial images from one subject and an accomplice, who could apply for a valid passport exploiting the accomplice's identity. Morphing takes place in the enrolment process stage. The threat of morphing attacks is known for border crossing or identification control scenarios. 
A morphing attack's success depends on the decision of human observers, especially a passport identification expert. The real-life application for a border police expert who compares the passport reference image of the traveller (digital extracted from the embedded chip) with the facial appearance of the traveller \cite{venkatesh2020face} is too hard because of the improvements of the morphing tools and because of the difficulty for the human expert to localise facial areas, in which morphing artefacts are present.

This work proposes to add an extra stage of feature selection after feature extraction based on Mutual Information $MI$ to estimate and keep the most relevant and remove the most redundant features from the face images to separate bona fide and morphed images. The high redundancy between features confuses the classifier.

The contributions of this work are described as follows: a) Identify the most relevant and less redundant features from faces that allow us to separate bona fide from morphed images. b) Localise the position of the most relevant areas on the images. c) Visualise the areas that contain morphing artefacts d) Reduce the algorithm's complexity, sending fewer features to the classifier. e) Analysis of the feature level fusion, the intensity, shape, and texture information. 
All these contributions may help to guide the manual inspection of morphed images.

This paper is organised as follows: a summary background in features selection and $MI$ is presented in section \ref{FS}. The relate work is describe in Section \ref{sec:related}. The methods are described in Section \ref{method}. The database are described in section \ref{database} and the experiments and results are presented in section \ref{exp} and conclusion are presented in section \ref{conclusion}.

\section{Related work}\label{sec:related}

Face morphing attack has captured the interest of the research community and government agencies in Europe. For instance the EU founded the iMARS project \footnote{\url{https://cordis.europa.eu/project/id/883356}}, developing new techniques of manipulation and detection of morphed images.

Ferrera et al. \cite{Ferrara} was the first to investigate the face recognition system's vulnerability with regards to morphing attacks. He has evaluated the feasibility of creating deceiving morphed face images and analysed the robustness of commercial face recognition systems in the presence of morphing.

Scherlag et al. \cite{Scherhag_survey} studied the literature and developed a survey about the impact of morphing images on face recognition systems. The same author \cite{ScherhagDeep} proposed a face representation from embedding vectors for differential morphing attack detection, creating a more realistic database, different scenarios, and constraints with four automatic morphed tools. He also reported detection performances for several texture descriptors in conjunction with machine learning techniques.

Indeed, the NIST FRVT MORPH \cite{Ngan2020FaceRV} evaluates and reports the performances of different morph detection algorithms organised in three tiers according to the morph images quality. Tier 1 evaluates low-quality morph images; Tier 2 considers automatic morph images; and Tier 3 for high-quality images. Further, the NIST report is organised w.r.t local (crop faces) and global (passport-photos) morphing algorithms. This fact confirms and shows that morphing images is a problem considering many scenarios. 

Most of the state-of-the-art approaches are using machine learning and deep learning to detect and classify morph images. Also, they are utilising embedding vectors from deep learning approaches to detect and classify the images. However, those approaches did not analyse the most relevant features and their localisation on the original images. An efficient feature selection method may help to improve this limitation. 

Regarding feature selection, in image understanding, raw input data often has very high dimensionality and a limited number of samples. In this area, feature selection plays an important role in improving accuracy, efficiency and scalability of the object identification process.  Since relevant features are often unknown a priori in the real world, irrelevant and redundant features may be introduced to represent the domain. 
However, using more features implies increasing computational cost in the feature extraction process, slowing down the classification process and  also increasing the time needed for training and validation, which may  lead to classification over-fitting. As is the case in most image analysis problems, with a limited amount of sample data, irrelevant features may obscure the distributions of the small  set of relevant features and confuse the classifiers.

Peng et al. \cite{Peng2005} develop a general framework to analyse the interaction between the redundancy and the relevance of the features in a machine learning method to look at the most valuable features based on $MI$.

Guyon et al.\cite{Guyon2006} proposed the Conditional Mutual Information Maximisation (CMIM) to estimate the relationship of the relevance of the features among three pairs of features.

Vergara et al. \cite{Vergara} proposed an improvement for CMIM \cite{Guyon2006} approach based on the selection of the first relevant feature. The traditional method maximised the conditional mutual information to select relevant features. This author proposes the average of the $MI$ to reduce the difference among chosen features.

Tapia et al. \cite{iriscode, cmimw} used the measures of $MI$ to guide the selection of bits from the iris code to be used as features in gender prediction. Also, in \cite{cmimw} used complementary information to create clusters of the most relevant features based on information theory to classify gender from faces. 

According to those previous works, we believed that $MI$ is suitable for detecting morphed images to localised and detect the artefact present in morphed images using an efficient number of features.

\section{Methods}
\label{method}
\begin{figure*}[]
\centering
	\includegraphics[scale=0.36]{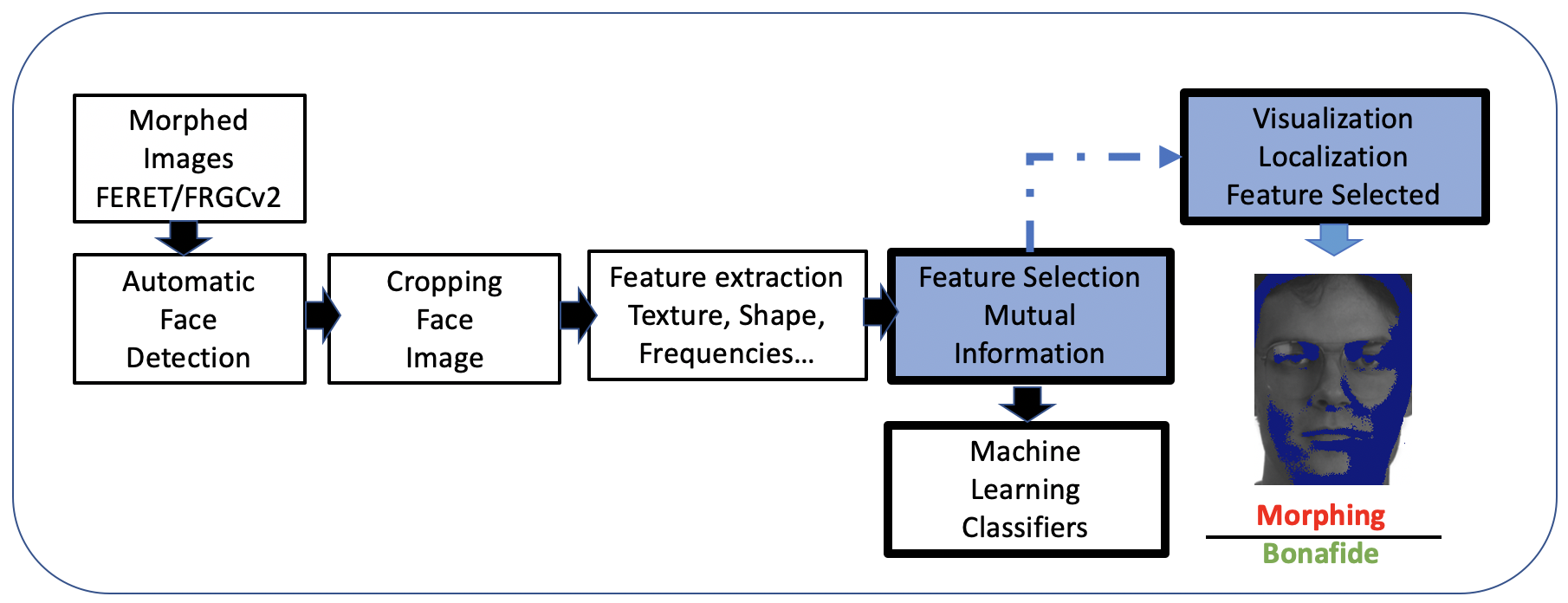}
\caption{Framework proposed with feature selection stage.}
\label{fig:framwork}
\end{figure*}

Figure \ref{fig:framwork} shows the proposed framework used in this paper, where a feature selection stage is added after traditional feature extraction approaches.

\subsection{Feature extraction}

Three different features were extracted from the morphing face images: Intensity, Texture and Shape. 

\subsubsection{Intensity}

For raw data the intensity of the values in grayscale were used and normalised between 0 and 1. 

\subsubsection{Uniform Local Binary Pattern}

For texture, the histogram of uniform local binary pattern were used \cite{LBPs}. LBP is a gray-scale texture operator which characterises the spatial structure of the local image texture. Given a central pixel in the image, a binary pattern number is computed by comparing its value with those of its neighbours. The original operator used a $3\times3$ windows size. LBP features were computed from relative pixels intensities in a neighbourhood, as is show in the following equation:

\begin{equation}
LBP_{P,R}(x,y)=\bigcup_{(x',y')\in N(x,y)}h(I(x,y),I(x',y'))\label{eq:LBP-1}
\end{equation}

where $N(x,y)$ is vicinity around $(x,y)$, $\cup$  is the concatenation operator, $P$ is number of neighbours and $R$ is the radius of the neighbourhood.

The uniform Local Binary Pattern (uLBP) was used as texture information. The uLBP was introduced, extending the original LBP operator to a circular neighbourhood with a different radius size and a small subset of LBP patterns selected. In this work we use, ‘U2’ which refers to a uniform pattern. LBP is called uniform when it contains at most 2 transitions from 0 to  1 or 1 to 0, which is considered to be a circular code. Thus, the number of patterns is reduced from 256 to 59 bins.

\begin{figure}[]
\centering
	\includegraphics[scale=0.38]{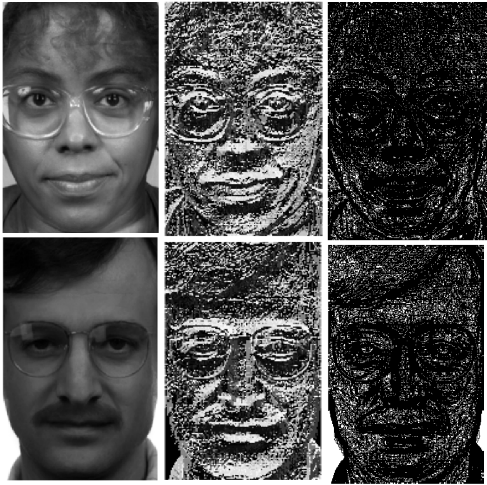}
\caption{Example of LBP images. Left:Grayscale image. Middle: traditional LBP (256 bins). Right: LBP with uniform pattern implementation (59 bins).}
\label{fig:lbp_images}
\end{figure}

The reasons for omitting the non-uniform patterns are twofold. First, most of the LBP in natural images are uniform. It was noticed experimentally 
that uniform patterns account for a bit less than 90\% of all patterns when using the (8,1) neighbourhood. In experiments with facial images, it was found that 90.6\% of the patterns in the (8,1) neighbourhood and 85.2\% of the patterns in the (8,2) neighbourhood are uniform \cite{ECCV_gender}.
The second reason for considering uniform patterns is the statistical robustness. Using uniform patterns instead of all the possible patterns has produced better recognition results in many applications. On one hand, there are indications that uniform patterns themselves are  more stable, i.e.  less prone to noise and on the other hand, considering only uniform patterns  makes the number of possible LBP labels significantly lower and reliable estimation of their distribution requires fewer samples. See Figure \ref{fig:lbp_images}.

\subsubsection{Inverse Histogram Oriented Gradient}

\begin{figure}[]
\centering
	\includegraphics[scale=0.36]{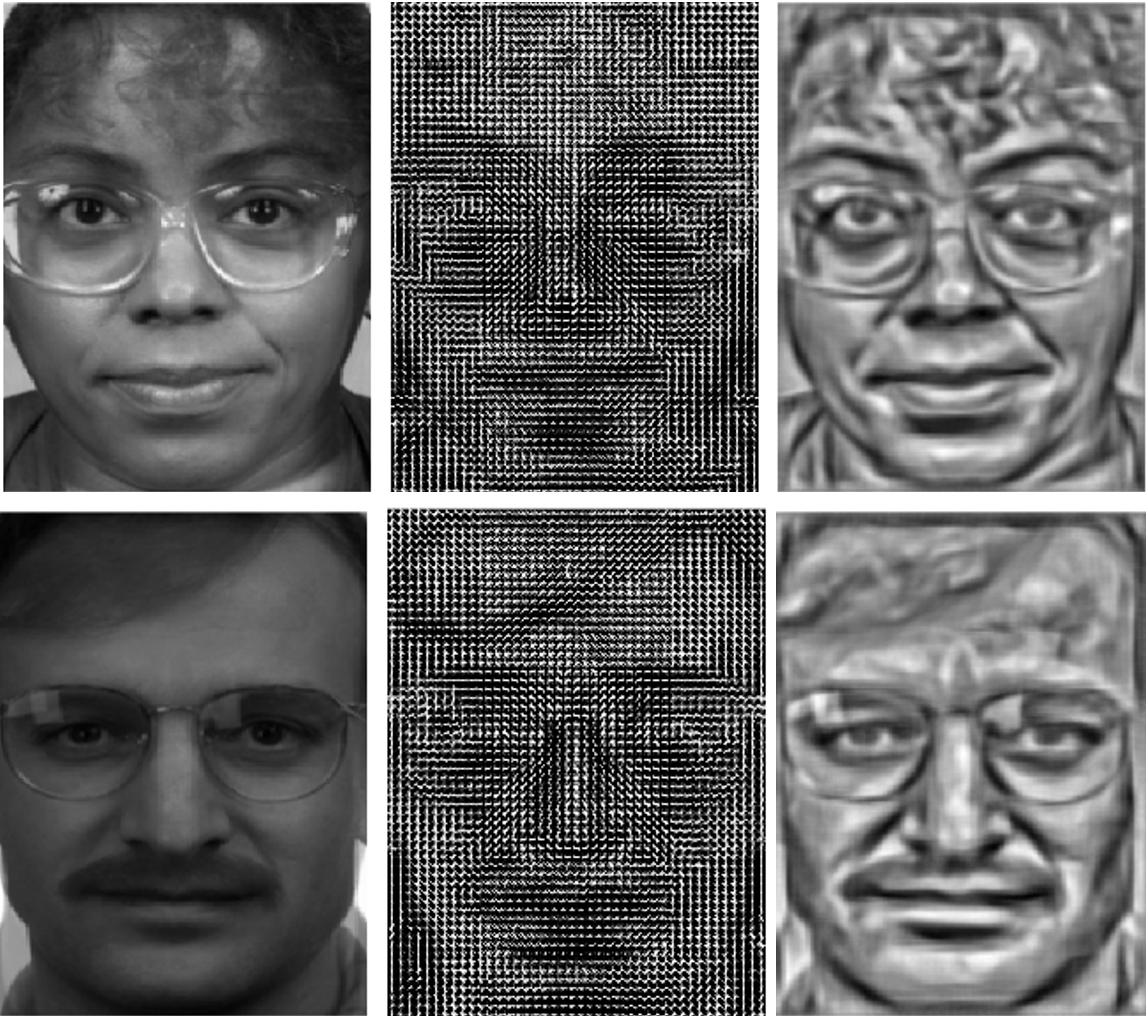}
\caption{Example images of inverse HOG. Left: Morphed images. Middle: Traditional HOG. Right: Inverse HOG.}
\label{fig:ihog}
\end{figure}

From Shape, the inverse Histogram of oriented gradients \cite{iHOG, HOG} were used. The Histogram of oriented gradient was proposed by Dalal et al. \cite{HOG}. The distribution directions of gradients (oriented gradients) are used as features. Gradients, $x$, and $y$ derivatives of an image are helpful because the magnitude of gradients is large around edges and corners (regions of abrupt intensity changes). We know that edges and corners contain more information about object shape than flat regions. However, this descriptor presents some problems. For instance, when we visualise the features for high-scoring false alarms in the object detection area,  they are wrong in image space. They look very similar to true positives in feature space. To avoid this limitation that confuses the classifiers, we used the visualisation proposed by Vondrik et al. \cite{iHOG} to select the best parameters that allows us to visualise the artefacts contained in morphed images. This implementation used $10\times12$ blocks and $3\times3$ filter sizes.
One example is shown in Figure \ref{fig:ihog}.

\subsection{Feature selection}
\label{FS}

\begin{figure}[H]
\centering
	\includegraphics[scale=0.40]{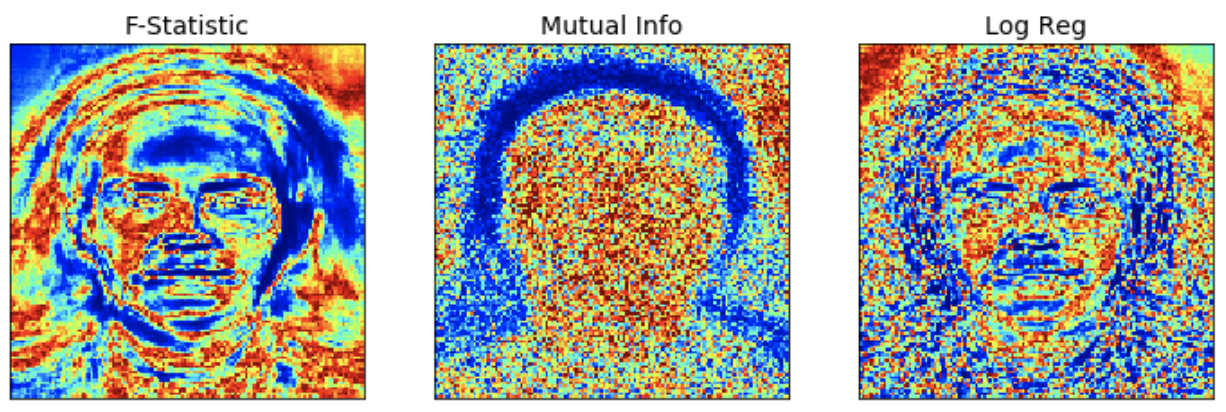}
\caption{Example images whit different correlation metrics. Red pixels represent the less correlated features.}
\label{fig:correlation}
\end{figure}

Feature selection (FS) is the process in which groups of features derived from image areas and textures respectively pixels (in raw images) from facial images out of a dataset are selected based on some measure or the correlation such as F-statistic, Logistic regression or $MI$ between the features and the class of the labels. See Figure \ref{fig:correlation}. 
It is closely related to feature extraction, a process in which feature vectors are created from the facial image. This takes place through domain transformation or manipulation of the data space and can be considered as selecting a subset of features.

Figure \ref{fig:correlation} shows a random morphed image with three different correlation metrics. The heat maps show the most correlated features in blue and the less correlated in red. All the features (relevant and redundant) are present in the image. 

FS can be classified into three main groups: Filters, Wrappers, and Embedding methods \cite{Guyon2006}. 
A filter does not have a dependency with classifiers when looking for the most relevant features as it. 
Filters estimates the correlation values according to the $MI$ values. Conversely, wrappers search for the most relevant features according to the classifier. Therefore, if the classifier changes, then the relevant features vary. The embedding method is looking to estimate an optimisation function according to the data and the classifier.

For this work, we propose to use a filter methods based on $MI$ as correlation metrics to estimate the most relevant features to classify bona fide versus morphed face images.

\subsection{Mutual information}

$MI$ is defined as a measure of how much information is contained jointly in two variables or how much information of one variable determines the other variable \cite{Cover1991}. $MI$ is the foundation for information theoretic feature selection since it provides a function for computing the relevance of a variable with respect to the target class \cite{Guyon2006}. The $MI$ between two variables, $x$ and $y$, is defined based on their joint probabilistic distribution $p(x,y)$ and the respective marginal probabilities $p(x)$ and $p(y)$ as: 

{
\begin{equation}
MI(x,y)=\sum_{{\scriptscriptstyle i,j}}{\textstyle p(x_{i},y_{j})log\frac{p(x_{i},y_{j})}{p(x_{i})p(y_{j})}}\label{eq:IM-1}.
\end{equation}
}{ \par}

A categorical $MI$ is used in this paper, which can be estimated by tallying the samples of categorical variables in the data building adaptive histograms to compute the joint probability distribution $p(x,y)$ and the marginal probabilities $p(x)$ and $p(y)$ based on the Fraser algorithm \cite{Fraser1986} for bona fide and morphing images. According to that, if more than two pairs of features reach the same value then, the information is \textbf{redundant}. Conversely, if a couple of features is not contained in any, other pair of features is considered \textbf{relevant} and therefore can help to disentangle and separate the two classes.
If a feature extracted from an image is randomly or uniformly distributed in different classes (bona fide or morph), then the $MI$ between these classes is zero. If a feature is strongly differently expressed for other classes (morph), it should have a large $MI$. Thus, we use $MI$ as a measure of the relevance of features presented in the images. 

The following protocol was used:

\begin{itemize}
    \item Each image of size $M \times N$ was flattened to $1\times M \times N$ for each class (bona fide and morphed).
    \item The matrix $A$ is formed by $K$ flattened images of size $1\times M \times N$ features, and the class vector (c).
    \item $MI$ for each pair of column of matrix $A$ is estimated.
    \item The relevance (Rl) and redundancy (Rd) are estimated from matrix $A$.
    \item The trade-off between the relevance and redundancy (Rl and Rd) matrices is estimated, sorted and indexes according to the $MI$ values.
    \item A vector $v$ with the index value of each column (feature) with the higher relevance and less redundant is formed.
    \item Only the $N$ columns according to with index value are selected.
    \item A small matrix from $A$ and element $v$ is conformed in the step of 100 features up to 1,000 features to be evaluated for the classifier.
\end{itemize}

Different implementations have been proposed in state-of-the-art \cite{Guyon2006} to estimate the trade-off between relevance and redundancy. Estimate all the combinations $2^N$ to remove all the redundancy is not possible because of high dimensionality problem. Then, the following methods based on $MI$ and Conditional $MI$ have been used and are described as follows:

\subsection{minimum Redundancy Maximal Relevance (mRMR)}

Two forms of combining relevance and redundancy operations are reported in \cite{Peng2005}; $MI$ difference $\left(MID\right)$, and $MI$ quotient $\left(MIQ\right)$. Thus, the $mRMR$ feature set is obtained by optimising $MID\:\:$ and $MIQ\:$ simultaneously. The trade-off both conditions requires to integrate them into a single criterion function \cite{Peng2005} as follows:

{
\scriptsize
\begin{equation}
f^{mRMR}(X_{i})=MI(c;fi)-\frac{1}{S}\sum MI(fi;fs),\label{eq:mrmr}
\end{equation}
}{ \par}

where,{\small{} $MI(c;fi)$} measures the relevance of the feature $f_i$ to be added for the class $c$, and the term{\small{} $\frac{1}{S}\sum_{fi\epsilon S}MI(fi;fs)$}
estimates the redundancy of the $fi_{th}$ feature with respect to the previously selected features $f_s$ to belong to set $S$.

\subsection{Normalised Mutual Information Feature Selection (NMIFS)}

Estevez et al.  \cite{Estevez2009} proposed with the Normalised Mutual Information (NMIFS) an improved version of mRMR based on the normalised feature of $MI$. The $MI$ between two random variables is bounded above by the minimum of their entropies $H$. As the entropy of a feature could vary greatly, this measure should be normalised before applying it to a global set of features as follows:

\begin{equation}
f^{NMIFS}(X_{i})=MI(c;fi)-\frac{1}{\textbar}S{\textbar}\sum_{fi\epsilon S}MI_{N}(fi;fs)\label{eq:nmifs}
\end{equation}

Where, $MI_{N}\:$ is the normalised $MI$ by the minimum entropy of both features, as defined in:

\begin{equation}
MI_{N}(fi;fs)=\frac{MI(fi;fs)}{min(H(fi),H(fs))}\label{eq:IN}
\end{equation}

\subsection{Conditional Maximisation Mutual Information (CMIM)}

The $CMIM$ criterion is a tri-variate measure of the information associated with a single feature about the class, conditioned upon an already selected feature \cite{Fleuret}. It loops over the chosen features and assigns each candidate to feature a score based upon the lowest Conditional Mutual Information $(CMI)$ between the features selected, the candidate feature, and the class \cite{Guyon2006, Fleuret}. Then, the selected feature is the one with the maximum score.

{
\scriptsize
\begin{equation}
CMIM=\begin{cases}
arg\: max_{fi\in F}\left\{ MI(fi;c) for\: S=\emptyset\right\}  \\ 
arg\: max_{fi\in F/S}\left\{ min_{fj\in S}\, MI(fi;c/fj)\right\}  \\ for\: S\neq\emptyset.
\end{cases}
\end{equation}
}{\footnotesize \par}

\subsection{Conditional Maximisation Mutual Information-2 (CMIM2)}
The $CMIM$ criterion selects relevant variables and avoids redundancy. However, it does not necessarily choose a variable that is complementary to the already chosen variables. A variable with high complementarity information (max) to the already selected variable will be had by a high $(CMI)$.
In general, in problems where the variables are highly complementary (or dependent) to predict $c$, the $CMIM$ algorithm will fail to find that dependence among the variables. The $CMIM-2$ \cite{Vergara} was proposed in order to improve $CMIM$ and changes the max function for the average function ($1/d$). Then, the selected feature is the one with the average score.

{
\scriptsize
\begin{equation}
MI(x,y)=1/d \sum_{f,j{ \in S}}{MI(f_i; c\mid f_j)}\label{eq:IM-1}.
\end{equation}
}{ \par}

\begin{figure*}[]
    \hfill
    \begin{minipage}{.165\linewidth}
    	\includegraphics[width=\linewidth]{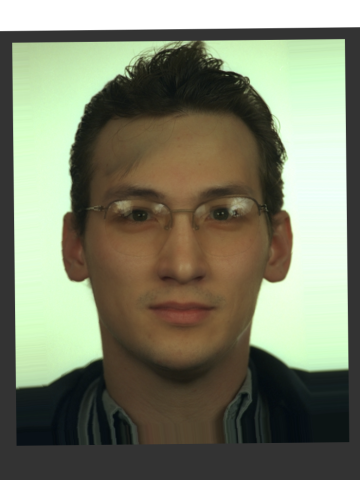}
    	\caption*{FERET Subject 1}
    \end{minipage}%
    \hfill
    \begin{minipage}{.165\linewidth}
    	\includegraphics[width=\linewidth]{images/Examples/FERET/ff_01208_940128_fa_a.png_vs_00192_940128_fa.png}
    	\caption*{FaceFusion}
    \end{minipage}%
    \hfill
    \begin{minipage}{.16\linewidth}
    	\includegraphics[width=\linewidth]{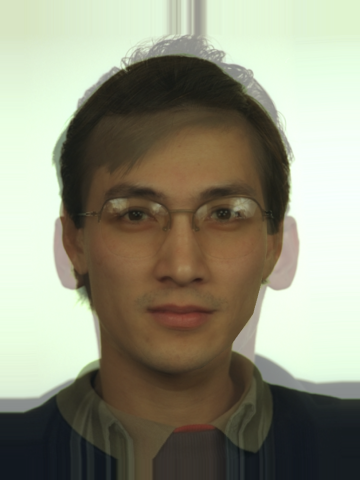}
    	\caption*{FaceMorpher}
    \end{minipage}%
    \hfill
    \begin{minipage}{.16\linewidth}
    	\includegraphics[width=\linewidth]{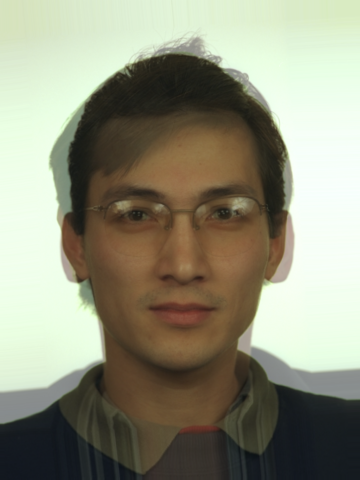}
    	\caption*{OpenCV Morpher}
    \end{minipage}%
    \hfill
        \begin{minipage}{.16\linewidth}
    	\includegraphics[width=\linewidth]{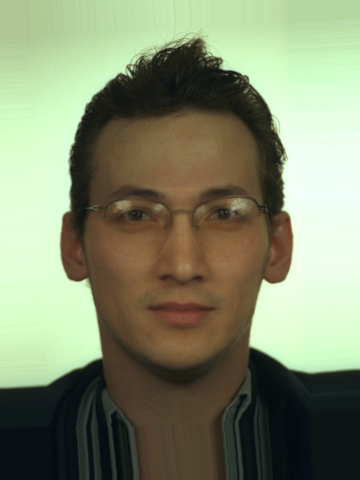}
    	\caption*{UBO-Morpher}
    \end{minipage}%
    \hfill
        \begin{minipage}{.16\linewidth}
    	\includegraphics[width=\linewidth]{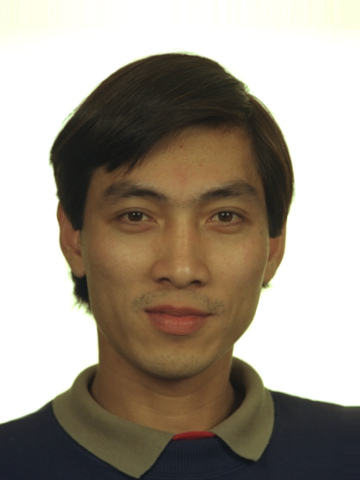}
    	\caption*{Subject2}
    \end{minipage}%
    \vspace{1ex}
    \hfill
    \begin{minipage}{.15\linewidth}
    	\includegraphics[width=\linewidth]{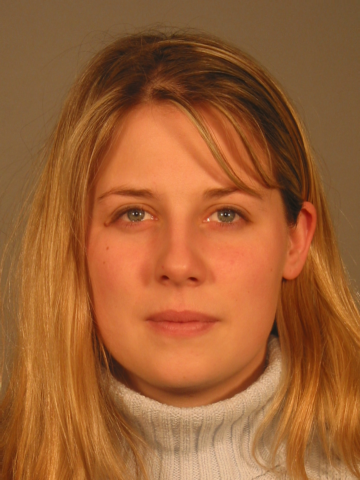}
    	\caption*{FRGCv2 Subject 1}
    \end{minipage}%
    \hfill
    \begin{minipage}{.17\linewidth}
    	\includegraphics[width=\linewidth]{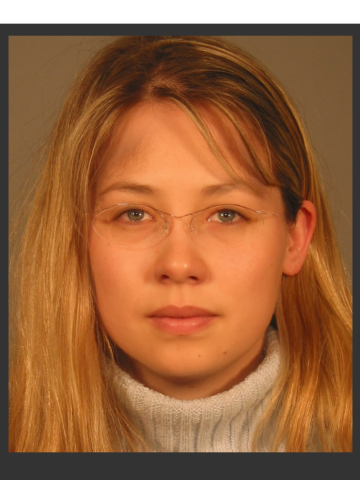}
    	\caption*{FaceFusion}
    \end{minipage}%
    \hfill
    \begin{minipage}{.16\linewidth}
    	\includegraphics[width=\linewidth]{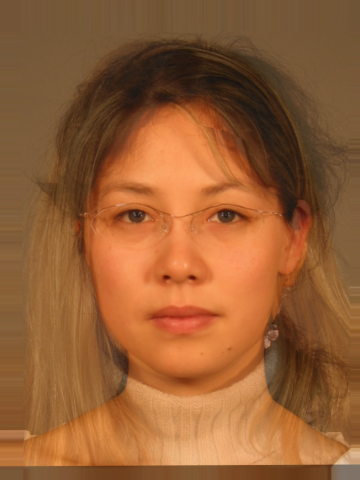}
    	\caption*{FaceMorpher}
    \end{minipage}%
    \hfill
    \begin{minipage}{.16\linewidth}
    	\includegraphics[width=\linewidth]{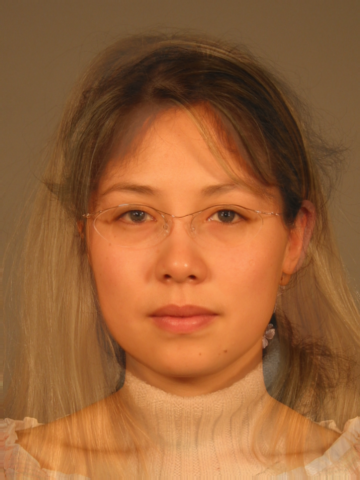}
    	\caption*{OpenCV Morpher}
    \end{minipage}%
    \hfill
    \begin{minipage}{.16\linewidth}
    	\includegraphics[width=\linewidth]{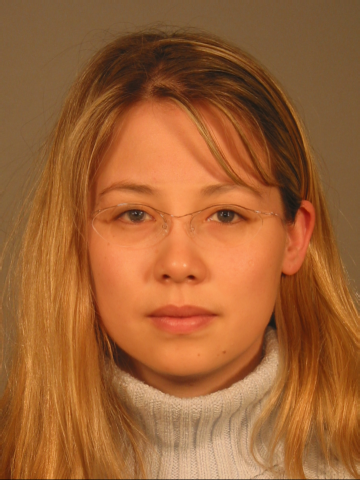}
    	\caption*{UBO-Morpher}
    \end{minipage}%
    \hfill
    \begin{minipage}{.16\linewidth}
    	\includegraphics[width=\linewidth]{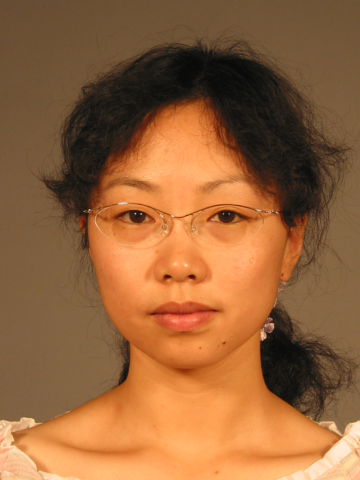}
    	\caption*{Subject 2}
    \end{minipage}%
    \hfill
\caption{Examples of different morphing algorithms for two subjects in the FERET and FRGCv2 databases}
\label{fig:datasetex}
\end{figure*}

\section{Databases}
\label{database}

The FERET and FRGCv2 databases were used to create the morph images based on the protocol described by \cite{ScherhagDeep}. A summary of the databases is presented in Table \ref{tab:database}. All the images were captured in a controlled scenario and include variations in pose and illumination.
FRGCv2 presents images more compliant to the passport portrait photo requirements. The images contain illumination variation, different sharpness and changes in the background. 
The original images have the size of $720\times960$ pixels. For this paper, the faces were detected, and images were resized and reduced to $180 \times 240$ pixels. These images still fulfill the resolution requirement of the intra-eye distance of 90 pixels defined by ICAO-9303-p9-2015. 
The $\alpha$ value to define the contribution of each subject to morph image results was 0.5.

Figure \ref{fig:datasetex} shows examples of the morphing portrait images and the different output qualities with the artefact in their background. For instances OpenCV implementation. 

\begin{table}[H]
\scriptsize
\centering
\caption{Number of images used for FERET and FRGCv2 Database. Column 1, show the software used to create morph images. The number of images is per dataset.}
\begin{tabular}{lllll}
\hline
\textbf{Database} & \textbf{Nº Subjects} & \textbf{Bona fide} & \textbf{Morphs} & \textbf{Probes} \\ \hline
FRGCv2            & 533              & 984               & 964             & 1726            \\ \hline
FERET             & 529              & 529               & 529             & 791             \\ \hline
FaceFusion        & 533              & 984               & 964             & 1726            \\ \hline
FaceMorpher       & 533              & 984               & 964             & 1726            \\ \hline
FaceOpenCV           & 533              & 984               & 964             & 1726            \\ \hline
UBO-Morpher       & 533              & 984               & 964             & 1726            \\ \hline
\end{tabular}
\label{tab:database}
\end{table}

The following algorithms were used to create morph images:
\begin{itemize}
    \item FaceFusion is a proprietary morphing algorithm, developing for IOS app \footnote{\url{www.wearemoment.com/FaceFusion/}}. This algorithm to create high-quality morph images without visible artifact.
    \item FaceMorpher is an open-source algorithm to create morph images \footnote{\url{github.com/alyssaq/face\_\morpher}}. This algorithm introduce also some artifacts in the background.
    \item FaceOpenCV, this algorithm is based on the OpenCV implementation. \footnote{\url{www.learnopencv.com/face-morph-using-opencv-cpp-python}}. The images contain visible artifacts in the background and some areas of the face.
    \item Face UBO-Morpher. The University of Bologna developed this algorithm. The resulting images are of high quality without artifact in the background. 
\end{itemize}

\begin{figure}[]
\centering
	\includegraphics[scale=0.14]{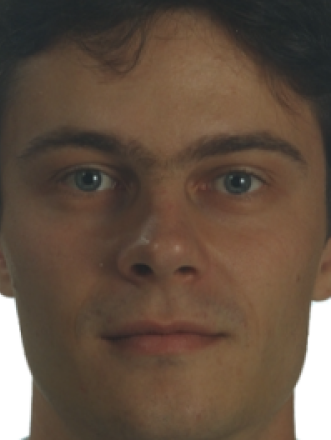}\includegraphics[scale=0.149]{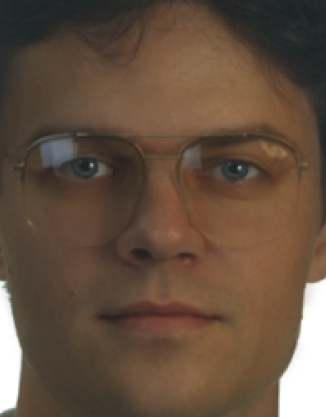}\includegraphics[scale=0.155]{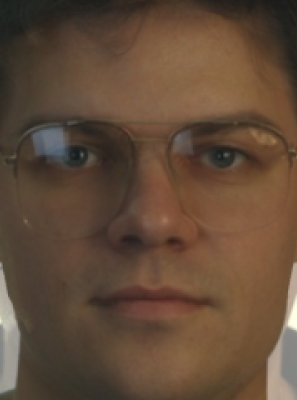}\includegraphics[scale=0.148]{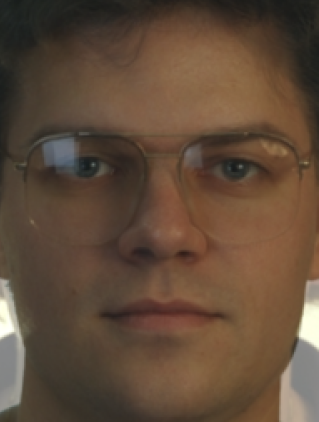}\includegraphics[scale=0.158]{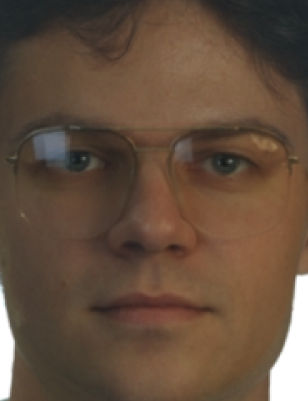}
\caption{Examples of FERET cropped images. From Left to Right: Bona fide, FaceFusion, FaceMorpher, OpenCV, UBO-Morpher implementations.}
\label{fig:morphing}
\end{figure}

As we mentioned before, after creation of the morphed images, all the faces were cropped using a modified dlib face detector implementation \footnote{https://www.pyimagesearch.com/2018/09/24/opencv-face-recognition/}. Figure \ref{fig:morphing} shows examples of the FERET cropped face database. We can observe that cropped images represent a more challenging scenario because all the background artefacts of the morphing process result were removed. However, some artefacts remain and can be observed in the images, for instances for the FaceMorpher and OpenCV implementations.

\section{Experiments and Results} 
\label{exp}

\begin{figure}[H]
\centering
	\includegraphics[scale=0.25]{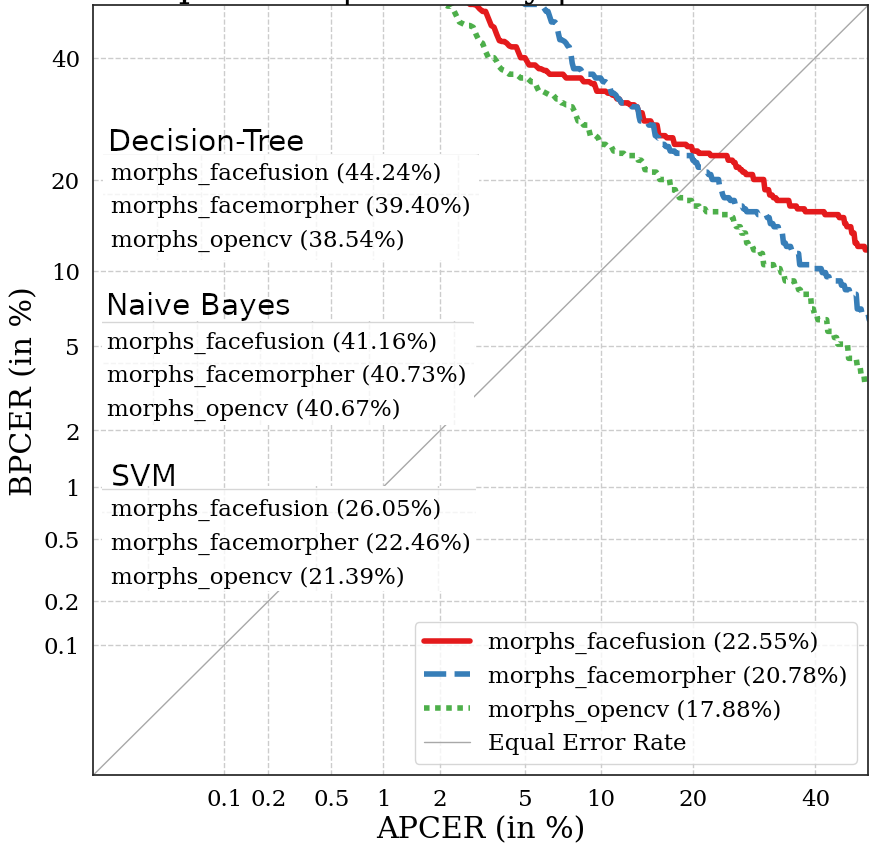}
\caption{DET Curves comparing the baseline classifiers using RF. RF and SVM reached the best results. KNN, LOGIT and MLP are not showed in the curve because of poor results. }
\label{fig:all_curves}
\end{figure}

This section presents the quantitative results of the proposed scheme based on feature selection for automated single-morph attack detection. In addition to the proposed system, we evaluated six different contemporary classifiers such as K-Nearest Neighbors (KNN), Logistic regression (LOGIT), Support Vector Machine (SVM), Decision Tree (DT), Random Forrest (RF), and Multilayer Perceptron (MLP). Overall, Random Forest and SVM reached the best results. See Figure \ref{fig:all_curves}. To compare and to estimate the baseline method, only the Random Forest classifier was used.  

The experiments, tested a leave-one-out (LOO) protocol and an RF classifier with 300 trees. These datasets allow subject-disjoint results to be computed; that is, no subject has an image in both the training and the testing subset. 

The FERET and FRGCv2 databases were partitioned to have 60\% training and 40\% testing data for feature selection. The selection of features was made using only the training set. The output of the four methods delivers the index of each column of the matrix $A$ that represents the more relevant features. The number of features were evaluated in steps of 100 features up to the end of the vector.

The performance of the detection algorithms is reported according to metrics defined in ISO/IEC 30107-3. The Attack Presentation Classification Error Rate (APCER) is defined as the proportion of attack presentations using the same attack instrument species incorrectly classified as bona fide in a specific scenario. The bona fide Presentation Classification Error Rate (BPCER) is defined as the proportion of bona fide images incorrectly classified as a morphing in the system. The D-EER is the operation point where APCER = BPCER is reported for the different morphing methods.

\begin{table*}[]
\centering
\scriptsize
\caption{Baseline performance reported in \% of D-EER for FERET LOO trained on FaceFusion and FaceMorpher.}
\label{tab:ff-fm_feret}
\begin{tabular}{|c|c|c|c|c|c|c|c|c|c|}
\hline
Train     & \multicolumn{4}{c|}{FACEFUSION}              & Train     & \multicolumn{4}{c|}{FACEMORPHER}            \\ \hline
Method    & FACEMORPHER & OpenCV & UBO-MORPHER & Average & Method    & FACEFUSION & OpenCV & UBO-MORPHER & Average \\ \hline
RAW       & 37.47       & 35.67  & 41.35       & 38.16   & RAW       & 49.23      & 23.53  & 49.6        & 40,79   \\ \hline
HOG       & 38.83       & 40.47  & 40.4        & 39.90   & HOG4      & 42.03      & 37.14  & 42.07       & 40.41   \\ \hline
LBP1      & 27.35       & 32.33  & 38.53       & 32.74   & LBP1      & 45.45      & 32.88  & 42.67       & 40.33   \\ \hline
LBP2      & 24.01       & 27.46  & 37.31       & 29.59   & LBP2      & 42.8       & 30.65  & 41.79       & 38.41   \\ \hline
LBP3      & 24.88       & 26.92  & 37.03       & 29.61   & LBP3      & 40.28      & 26,55  & 40,45       & 35,76   \\ \hline
LBP4      & 23.25       & 24.32  & 36.24       & 27.94   & LBP4      & 38.76      & 25.29  & 40.91       & 34.99   \\ \hline
LBP5      & 24.55       & 26.25  & 38.7        & 29.83   & LBP5      & 36,14      & 30.14  & 38.85       & 35.04   \\ \hline
LBP6      & 25.79       & 26.98  & 38.95       & 30.57   & LBP6      & 35.71      & 27.49  & 40.27       & 34.49   \\ \hline
LBP7      & 27.78       & 28.37  & 40.42       & 32.19   & LBP7      & 37.72      & 26,43  & 42,2        & 35.45   \\ \hline
LBP8      & 28.88       & 27.73  & 42.47       & 33.03   & LBP8      & 38.01      & 27.21  & 43.59       & 36.27   \\ \hline
\textbf{uLBP\_ALL}  & 23.71       & 26.34  & 38.02       & \textbf{29.36}   & \textbf{uLBP\_ALL}  & 38.59      & 30.05  & 40.33       & \textbf{36.32}   \\ \hline
LBP\_VERT & 26.76       & 30.1   & 23.98       & 26.95   & LBP\_VERT & 40.99      & 24.81  & 42.68       & 36.16   \\ \hline
LBP\_HOR  & 26.59       & 28.66  & 37.69       & 30.98   & LBP\_HOR  & 41.95      & 25,06  & 42,71       & 36,57   \\ \hline
FUSION    & 43.64       & 44.56  & 46.88       & 45.03   & FUSION    & 44.28      & 32.31  & 46.68       & 41.09   \\ \hline
\end{tabular}
\end{table*}

\begin{table*}[]
\centering
\scriptsize
\caption{Baseline performance reported in \% of D-EER for FERET LOO trained on OpenCV and UBO-Morpher.}
\label{tab:ocv-ubo_feret}
\begin{tabular}{|c|c|c|c|c|c|c|c|c|c|}
\hline
Train     & \multicolumn{4}{c|}{OpenCV}                      & Train     & \multicolumn{4}{c|}{UBO-MORPHER}                 \\ \hline
Method    & FACEFUSION & FACEMORPHER & UBO-MORPHER & Average & Method    & FACEFUSION & FACEMORPHER & UBO-MORPHER & Average \\ \hline
RAW       & 47.45      & 20.21       & 48.82       & 38.83   & RAW       & 35.55      & 40.45       & 37.46       & 37.82   \\ \hline
HOG       & 43.17      & 35.7        & 40.032      & 39.63   & HOG       & 39.77      & 35.74       & 35.82       & 37.11   \\ \hline
LBP1      & 44.6       & 25.72       & 41.68       & 37.33   & LBP1      & 42.6       & 27.45       & 32.93       & 34.33   \\ \hline
LBP2      & 41.28      & 24.85       & 40.28       & 35.47   & LBP2      & 40.28      & 25.26       & 29.86       & 31.80   \\ \hline
LBP3      & 37.66      & 23.95       & 39.52       & 33.71   & LBP3      & 35.99      & 24.97       & 25.58       & 28.85   \\ \hline
LBP4      & 36.08      & 22.72       & 38.52       & 32.44   & LBP4      & 36.03      & 24.33       & 26.99       & 29.12   \\ \hline
LBP5      & 35.76      & 25.56       & 38.5        & 33.27   & LBP5      & 34.31      & 26.94       & 28.64       & 29.96   \\ \hline
LBP6      & 37.03      & 28.74       & 40.58       & 35.45   & LBP6      & 37.74      & 30.36       & 30.21       & 32.77   \\ \hline
LBP7      & 37.74      & 25.51       & 42.47       & 35.24   & LBP7      & 38.56      & 32.08       & 31.5        & 34.05   \\ \hline
LBP8      & 37.66      & 27.47       & 42.86       & 36.00   & LBP8      & 39.79      & 34.58       & 32.16       & 35.51   \\ \hline
\textbf{uLBP\_ALL}  & 42.23      & 22.48       & 41.36       & \textbf{35.36}   & \textbf{uLBP\_ALL}  & 41.58      & 25.78       & 29.81       & \textbf{32.39}   \\ \hline
LBP\_VERT & 40.6       & 24.77       & 42.63       & 36.00   & LBP\_VERT & 38.51      & 29.6        & 31.26       & 33.12   \\ \hline
LBP\_HOR  & 41.5       & 23.4        & 42.45       & 35.78   & LBP\_HOR  & 38.66      & 29.55       & 31.7        & 33.30   \\ \hline
FUSION    & 44.85      & 28.84       & 45.92       & 39.87   & FUSION    & 46.17      & 43.57       & 44.05       & 44.60   \\ \hline
\end{tabular}
\end{table*}
\begin{table*}[]
\scriptsize
\centering
\caption{Baseline performance reported in \% of D-EER for FRGCv2 LOO trained on FaceMorpher and FaceFusion.}
\label{tab:frgc_fm-ff}
\begin{tabular}{|c|c|c|c|c|c|c|c|c|c|}
\hline
Train     & \multicolumn{4}{c|}{FACEFUSION}              & Train     & \multicolumn{4}{c|}{FACEMORPHER}            \\ \hline
Method    & FACEMORPHER & OpenCV & UBO-MORPHER & Average &           & FACEFUSION & OpenCV & UBO-MORPHER & Average \\ \hline
RAW       & 25.1        & 23.92  & 27.91       & 25.64   & RAW       & 41.71      & 13.97  & 42.3        & 32.66   \\ \hline
HOG      & 26.4        & 27.02  & 29.89       & 27.77   & HOG4      & 30.93      & 24.03  & 32.38       & 29.11   \\ \hline
LBP1      & 17.61       & 10.8   & 17.44       & 15.28   & LBP1      & 22.6       & 9.57   & 19.77       & 17.31   \\ \hline
LBP2      & 14.18       & 11.4   & 19.2        & 14.93   & LBP2      & 20.97      & 13.13  & 19.49       & 17.86   \\ \hline
LBP3      & 10.58       & 13.36  & 21.67       & 15.20   & LBP3      & 20.46      & 9.53   & 20.9        & 16.96   \\ \hline
LBP4      & 11.71       & 13.58  & 22.88       & 16.06   & LBP4      & 20.34      & 10.32  & 23          & 17.89   \\ \hline
LBP5      & 13.43       & 14.16  & 25.25       & 17.61   & LBP5      & 20.73      & 10.69  & 25.43       & 18.95   \\ \hline
LBP6      & 14.61       & 15.43  & 28.87       & 19.64   & LBP6      & 21.38      & 11.37  & 26.74       & 19.83   \\ \hline
LBP7      & 15.88       & 15.78  & 26.2        & 19.29   & LBP7      & 21.91      & 11.01  & 26.7        & 19.87   \\ \hline
LBP8      & 15.96       & 16.29  & 26.06       & 19.44   & LBP8      & 24         & 12.22  & 27.42       & 21.21   \\ \hline
\textbf{uLBP\_ALL}  & 10.05       & 12.38  & 20.36       & \textbf{14.26}   & \textbf{uLBP\_ALL}  & 22.44      & 7.99   & 21.64       & \textbf{17.36}   \\ \hline
LBP\_VERT & 13.81       & 14.43  & 20.9        & 16.38   & LBP\_VERT & 20.96      & 11.22  & 22.41       & 18.20   \\ \hline
LBP\_HOR  & 13.45       & 13.74  & 19.85       & 15.68   & LBP\_HOR  & 20.57      & 11     & 21.1        & 17.56   \\ \hline
FUSION   & 13.4        & 16.09  & 27.68       & 19.06   & FUSION2   & 27.79      & 15.21  & 26.81       & 23.27   \\ \hline
\end{tabular}
\end{table*}

\begin{table*}[]
\scriptsize
\centering
\caption{Baseline performance reported in \% of D-EER for FRGC LOO trained on OpenCV and UBO-Morpher.}
\label{tab:frgc_ocv_ubo}
\begin{tabular}{|c|c|c|c|c|c|c|c|c|c|}
\hline
Train     & \multicolumn{4}{c|}{OpenCV}                      & Train     & \multicolumn{4}{c|}{UBO-MORPHER}            \\ \hline
Method    & FACEFUSION & FACEMORPHER & UBO-MORPHER & Average & Method    & FACEFUSION & FACEMORPHER & OpenCV & Average \\ \hline
RAW       & 40.3       & 13.07       & 41.3        & 31.56   & RAW       & 22.58      & 23.83       & 22.24  & 22.88   \\ \hline
HOG       & 31.21      & 22.17       & 32.44       & 28.61   & HOG       & 27.44      & 25.38       & 26.5   & 26.44   \\ \hline
LBP1      & 20.94      & 13.29       & 17.88       & 17.37   & LBP1      & 20.54      & 6.32        & 9.92   & 12.26   \\ \hline
LBP2      & 20.52      & 8.43        & 18.89       & 15.95   & LBP2      & 20.18      & 7.38        & 9.86   & 12.47   \\ \hline
LBP3      & 19.57      & 7.26        & 20.11       & 15.65   & LBP3      & 19.59      & 9.54        & 11.76  & 13.63   \\ \hline
LBP4      & 20.76      & 8.05        & 23.1        & 17.30   & LBP4      & 18.99      & 10.69       & 12.65  & 14.11   \\ \hline
LBP5      & 20.4       & 9.1         & 24.63       & 18.04   & LBP5      & 20.38      & 13.28       & 13.64  & 15.77   \\ \hline
LBP6      & 21.66      & 10.66       & 26.57       & 19.63   & LBP6      & 21.19      & 15.7        & 15.82  & 17.57   \\ \hline
LBP7      & 22.64      & 10.68       & 26.84       & 20.05   & LBP7      & 20.26      & 16.85       & 16.47  & 17.86   \\ \hline
LBP8      & 23.5       & 11.94       & 27.78       & 21.07   & LBP8      & 22.71      & 19.37       & 19.52  & 20.53   \\ \hline
\textbf{uLBP\_ALL}  & 21.55      & 5.79        & 21.49       & \textbf{16.28}   & \textbf{uLBP\_ALL}  & 18.2       & 7.51        & 9.52   & \textbf{11.74}   \\ \hline
LBP\_VERT & 21.22      & 9.45        & 22.33       & 17.67   & LBP\_VERT & 18.19      & 13.77       & 14.07  & 15.34   \\ \hline
LBP\_HOR  & 20.98      & 9.62        & 21.97       & 17.52   & LBP\_HOR  & 18.34      & 13.69       & 13.9   & 15.31   \\ \hline
FUSION    & 27.22      & 11.7        & 26.31       & 21.74   & FUSION    & 27.59      & 11.24       & 13.22  & 17.35   \\ \hline
\end{tabular}
\end{table*}

\subsection{Experiment 1}

Three different kinds of features were extracted from faces. Intensity, HOG, and uLBP. From raw images, we used the values of intensity of the pixels normalised between 0 and 1. For shape, we used the histogram of HOG. For texture, the histogram of the Uniform Local Binary Patterns (uLBP) was used. For the uLBP all radii values were explored from uLBP81 to uLBP88. The fusion of LBPs was also investigated, concatenating the LBP81 up to LBP88 (LBP\_ALL). The vertical (uLBP\_VERT) and horizontal (uLBP\_HOR) concatenation of the image divided into 8 patches also was explored.
After feature extraction, we fused that information at the feature level by concatenating the feature vectors from different sources (Intensity, HOG, and uLBP) into a single feature vector that becomes the input to the classifier (FUSION). The classifier was trained with each feature extraction method's selected features and the fused chosen features. 

Table \ref{tab:ff-fm_feret} and \ref{tab:ocv-ubo_feret} show the baseline results for the intensity, shape and texture feature extraction methods. This baseline was estimated using a leave-one-out protocol for all the morphing methods. The intensity (Raw) and HOG reached the higher D-EER (worst results). Most of the time, the (LBP\_ALL) obtained the lower average D-EER results (Best results).

Table \ref{tab:ff-fm_feret} shows the results on the left side for the FERET database were trained with FaceFusion and tested with FaceMorpher, OpenCV, and UBO-Morpher. Right side, trained with FaceMorpher and tested with FaceFusion, OpenCV, and UBO-Morpher. 

Table \ref{tab:ocv-ubo_feret} shows the results on the left side for FERET database were trained with OpenCV and tested with FaceFusion, FaceMorpher, and UBO-Morpher. Right side, trained with UBO-Morpher and tested with FaceFusion, FaceMorpher and OpenCV. The same protocol was applied to Tables \ref{tab:frgc_fm-ff} and \ref{tab:frgc_ocv_ubo} with FRGCv2 database.

\subsection{Experiment 2}

This experiment explores the application of the proposed method based on feature selection. The four feature selection methods, mRMR, NMIFS, CMIM, and CMIM2, were applied in order to reduce the size of the data and estimate the position of the relevant features before entering classifiers from Intensity, HOG, and uLBP.
The best 5,000 from 43,200 features were extracted from the raw data (intensity). The best 1,000 from 1,048 features were extracted from HOG, and the best 400 features from 472 were selected from the fusion of uLBP (uLBP\_ALL). 

Table \ref{tab:hog_fea_feret} and \ref{tab:hog_fea_frgc} show the results for FERET and FRGCv2 database for single morphed detection from the best feature selected from \textbf{HOG} applied to FaceFusion, FaceMorpher, OpenCV-Morph and UBO-Morpher. The results reported shown an improved in comparison to the baseline in Experiment 1 using the HOG features extracted of the images. The number of feature was reduced on average down to 10\%. This reduction would enable the application in mobile devices hardware and also allow us to see the localisation of the most relevant features.  

\begin{table}[H]
\scriptsize
\caption{D-EER in \% of HOG + Fea / FERET. The figures in parenthesis represent the best number of features for each method.}
\label{tab:hog_fea_feret}
\begin{tabular}{|c|c|c|c|c|}
\hline
      & \begin{tabular}[c]{@{}c@{}}FaceFusion\\ (bestFea)\end{tabular} & \begin{tabular}[c]{@{}c@{}}FaceMorpher\\ (bestFea)\end{tabular} & \begin{tabular}[c]{@{}c@{}}OpenCV-Morph\\ (bestFea)\end{tabular} & \begin{tabular}[c]{@{}c@{}}UBO-Morpher\\ (bestFea)\end{tabular} \\ \hline
mRMR  & \begin{tabular}[c]{@{}c@{}}17.15\\ (400)\end{tabular}          & \begin{tabular}[c]{@{}c@{}}7.07\\ (700)\end{tabular}           & \begin{tabular}[c]{@{}c@{}}6.15\\ (700)\end{tabular}            & \begin{tabular}[c]{@{}c@{}}15.68\\ (100)\end{tabular}           \\ \hline
NMIFS & \begin{tabular}[c]{@{}c@{}}19.98\\ (300)\end{tabular}          & \begin{tabular}[c]{@{}c@{}}9.74\\ (300)\end{tabular}           & \begin{tabular}[c]{@{}c@{}}5.84\\ (800)\end{tabular}            & \begin{tabular}[c]{@{}c@{}}13.88\\ (300)\end{tabular}           \\ \hline
CMIM  & \begin{tabular}[c]{@{}c@{}}17.83\\ (300)\end{tabular}          & \begin{tabular}[c]{@{}c@{}}5.84\\(600)\end{tabular}                                                       &\begin{tabular}[c]{@{}c@{}} 7.07\\(500)\end{tabular}                                                       & \begin{tabular}[c]{@{}c@{}}11.07\\ (400)\end{tabular}           \\ \hline
CMIM2 & \begin{tabular}[c]{@{}c@{}}8.12\\ (300)\end{tabular}          & \begin{tabular}[c]{@{}c@{}}4.92\\ (900)\end{tabular}           & \begin{tabular}[c]{@{}c@{}}6.15\\ (500)\end{tabular}            & \begin{tabular}[c]{@{}c@{}}13.52\\ (300)\end{tabular}           \\ \hline
\end{tabular}
\end{table}

\begin{table}[H]
\scriptsize
\caption{D-EER in \% of HOG + Fea / FRGCv2. The figures in parenthesis represent the best number of features for each method.}
\label{tab:hog_fea_frgc}
\begin{tabular}{|c|c|c|c|c|}
\hline
      & \begin{tabular}[c]{@{}c@{}}FaceFusion\\ (bestFea)\end{tabular} & \begin{tabular}[c]{@{}c@{}}FaceMorpher\\ (bestFea)\end{tabular} & \begin{tabular}[c]{@{}c@{}}OpenCV-Morph\\ (bestFea)\end{tabular} & \begin{tabular}[c]{@{}c@{}}UBO-Morpher\\ (bestFea)\end{tabular} \\ \hline
mRMR  & \begin{tabular}[c]{@{}c@{}}6.83\\ (1000)\end{tabular}          & \begin{tabular}[c]{@{}c@{}}15.06\\ (900)\end{tabular}           & \begin{tabular}[c]{@{}c@{}}2.17\\ (1000)\end{tabular}            & \begin{tabular}[c]{@{}c@{}}4.99\\ (500)\end{tabular}           \\ \hline
NMIFS & \begin{tabular}[c]{@{}c@{}}4.83\\ (600)\end{tabular}          & \begin{tabular}[c]{@{}c@{}}2.1\\ (900)\end{tabular}           & \begin{tabular}[c]{@{}c@{}}1.68\\ (900)\end{tabular}            & \begin{tabular}[c]{@{}c@{}}4.73\\ (700)\end{tabular}           \\ \hline
CMIM  & \begin{tabular}[c]{@{}c@{}}6.50\\ (700)\end{tabular}          & \begin{tabular}[c]{@{}c@{}}1.83\\ (400)\end{tabular}           & \begin{tabular}[c]{@{}c@{}}2.17\\ (1000)\end{tabular}            & \begin{tabular}[c]{@{}c@{}}3.83\\ (500)\end{tabular}           \\ \hline
CMIM2 & \begin{tabular}[c]{@{}c@{}}6.65\\ (900)\end{tabular}          & \begin{tabular}[c]{@{}c@{}}1.92\\ (900)\end{tabular}           & \begin{tabular}[c]{@{}c@{}}1.50 \\ (1000)\end{tabular}          & \begin{tabular}[c]{@{}c@{}}4.02\\ (600)\end{tabular}           \\ \hline
\end{tabular}
\end{table}

Table \ref{tab:uLBP_fea_feret} and \ref{tab:frgc_fea_ulbp} show the results for FERET and FRGCv2 database for single morphed detection from the best feature selected from the \textbf{fusion of uLBP} (LBP8,1 up to LBP 8,8) applied to FaceFusion, FaceMorpher, OpenCV-Morph and UBO-Morpher. The results reported shown an improved in comparison to the Experiment 1 using all the features extracted of the images. The number of feature also is reduced on average down to 10\% for texture features. 

\begin{table}[]
\scriptsize
\caption{D-EER in \% of Fusion uLBP + Fea / FERET. The figures in parenthesis represent the best number of features for each method.}
\label{tab:uLBP_fea_feret}
\begin{tabular}{|c|c|c|c|c|}
\hline
      & \begin{tabular}[c]{@{}c@{}}FaceFusion\\ (bestFea)\end{tabular} & \begin{tabular}[c]{@{}c@{}}FaceMorpher\\ (bestFea)\end{tabular} & \begin{tabular}[c]{@{}c@{}}OpenCV-Morph\\ (bestFea)\end{tabular} & \begin{tabular}[c]{@{}c@{}}UBO-Morpher\\ (bestFea)\end{tabular} \\ \hline
mRMR  & \begin{tabular}[c]{@{}c@{}}22.7\\ (200)\end{tabular}          & \begin{tabular}[c]{@{}c@{}}12.92\\ (400)\end{tabular}           & \begin{tabular}[c]{@{}c@{}}13.84\\ (400)\end{tabular}            & \begin{tabular}[c]{@{}c@{}}23.37\\ (200)\end{tabular}           \\ \hline
NMIFS & \begin{tabular}[c]{@{}c@{}}21.22\\ (100)\end{tabular}          & \begin{tabular}[c]{@{}c@{}}11.84\\ (300)\end{tabular}           & \begin{tabular}[c]{@{}c@{}}12.30\\ (400)\end{tabular}            & \begin{tabular}[c]{@{}c@{}}23,98\\ (200)\end{tabular}           \\ \hline
CMIM  & \begin{tabular}[c]{@{}c@{}}21.53\\ (100)\end{tabular}          & \begin{tabular}[c]{@{}c@{}}12.30\\ (200)\end{tabular}           & \begin{tabular}[c]{@{}c@{}}11.84\\ (400)\end{tabular}            & \begin{tabular}[c]{@{}c@{}}22.75\\ (200)\end{tabular}           \\ \hline
CMIM2 & \begin{tabular}[c]{@{}c@{}}18.45\\ (100)\end{tabular}          & \begin{tabular}[c]{@{}c@{}}10.45\\ (400)\end{tabular}           & \begin{tabular}[c]{@{}c@{}}11.68\\ (400)\end{tabular}            & \begin{tabular}[c]{@{}c@{}}12.11\\ (200)\end{tabular}           \\ \hline
\end{tabular}
\end{table}

\begin{table}[]
\scriptsize
\caption{ D-EER  in \% of Fusion uLBP + Fea / FRGCv2. The figures in parenthesis represent the best number of features for each method.}
\label{tab:frgc_fea_ulbp}
\begin{tabular}{|c|c|c|c|c|}
\hline
      & \begin{tabular}[c]{@{}c@{}}FaceFusion\\ (bestFea)\end{tabular} & \begin{tabular}[c]{@{}c@{}}FaceMorpher\\ (bestFea)\end{tabular} & \begin{tabular}[c]{@{}c@{}}OpenCV-Morph\\ (bestFea)\end{tabular} & \begin{tabular}[c]{@{}c@{}}UBO-Morpher\\ (bestFea)\end{tabular} \\ \hline
mRMR  & \begin{tabular}[c]{@{}c@{}}9.15 \\ (400)\end{tabular}         & \begin{tabular}[c]{@{}c@{}}1.60\\ (200)\end{tabular}           & \begin{tabular}[c]{@{}c@{}}8.54\\ (300)\end{tabular}            & \begin{tabular}[c]{@{}c@{}}10.47\\ (300)\end{tabular}           \\ \hline
NMIFS & \begin{tabular}[c]{@{}c@{}}9.65\\ (400)\end{tabular}         & \begin{tabular}[c]{@{}c@{}}1.49\\ (200\end{tabular}            & \begin{tabular}[c]{@{}c@{}}4.50\\ (400)\end{tabular}            & \begin{tabular}[c]{@{}c@{}}8.99\\ (400)\end{tabular}           \\ \hline
CMIM  & \begin{tabular}[c]{@{}c@{}}8.52\\ (300)\end{tabular}          & \begin{tabular}[c]{@{}c@{}}1.09\\ (100)\end{tabular}           & \begin{tabular}[c]{@{}c@{}}4.17\\ (200)\end{tabular}            & \begin{tabular}[c]{@{}c@{}}8.70\\ (300)\end{tabular}           \\ \hline
CMIM2 & \begin{tabular}[c]{@{}c@{}}7.53\\ (300)\end{tabular}          & \begin{tabular}[c]{@{}c@{}}1,33\\ (200)\end{tabular}           & \begin{tabular}[c]{@{}c@{}}3.99\\ (400)\end{tabular}            & \begin{tabular}[c]{@{}c@{}}8.15\\ (400)\end{tabular}           \\ \hline
\end{tabular}
\end{table}

\begin{figure}[H]
\centering
\caption{\label{DET1-raw} FRGCv2 Feature selection for intensity features. X axis represents the number of the best features. Y axis represents the Accuracy in \%.}
\includegraphics[scale=0.30]{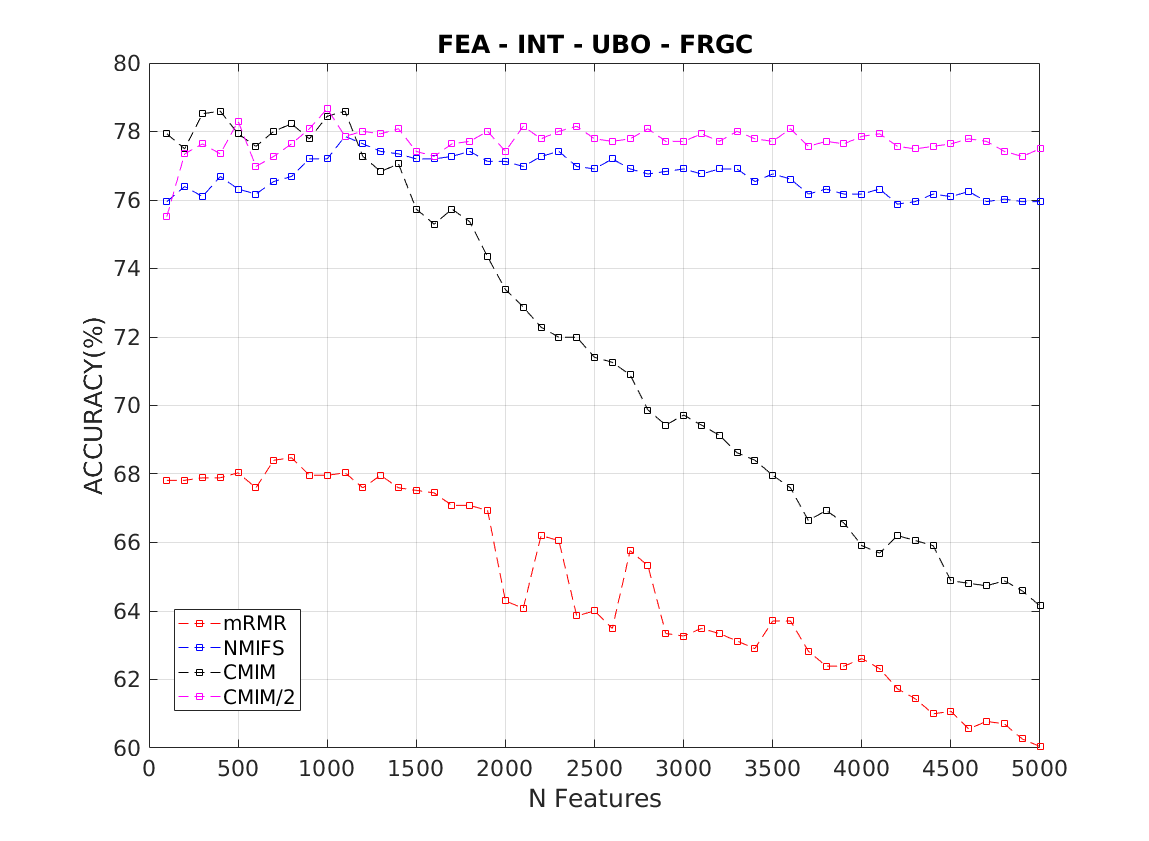}
\end{figure}

Figure \ref{DET1-raw} shows the accuracy obtained for the UBO-Morpher tool when features selected were applied from \textbf{intensity} features. The UBO-Morpher constitutes a high-quality morphing implementation and then is used and analysed on FERET and FRGCv2 databases. Conversely, FaceMorpher is the more straightforward method to be detected based on the artefacts present in the images.
The mRMR and NMIFS methods based on $MI$ obtained the lower results. The method based on conditional $MI$ (CMIM and CMIM-2) reached the best results. These results show that the complementary information captures the relationship between the feature selected and the feature candidate in a better way. CMIM with only 400 features and CMIM-2 with 1,000 features reached higher accuracy and lower D-EER.

\begin{figure}[]
\centering
\caption{\label{DET1-hog} FRGCv2 Feature selection for HOG features. X axis represents the number of the best features. Y axis represents the Accuracy in \%. }
\includegraphics[scale=0.38]{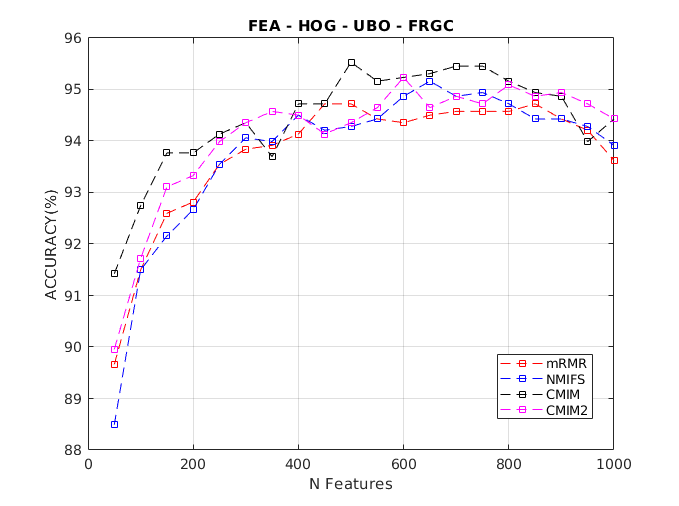}
\end{figure}

Figure \ref{DET1-hog} shows the accuracy obtained for the UBO-Morpher tool, when feature selected were applied from \textbf{HOG} features. Again, The method mRMR and NMIFS based on $MI$ obtained the lower results. The method based on conditional $MI$ (CMIM and CMIM-2) reached the best results with 500 and 600 features respectively. 

\begin{figure}[]
\centering
\caption{\label{DET1-lbp} FRGCv2 Feature selection for uLBP\_All features. X axis represents the number of the best features. Y axis represents the Accuracy in \%.}
\includegraphics[scale=0.47]{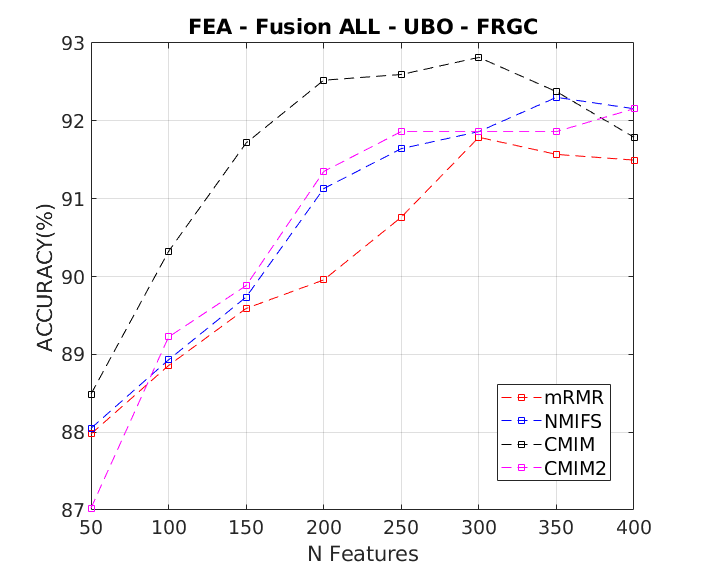}
\end{figure}

Figure \ref{DET1-lbp} shows the accuracy obtained for the UBO-Morpher tool, when feature selected were applied from the \textbf{fusion of uLBP}. This time NMIFS and CMIM reached the best results with 300 and 400 features respectively. Consolidating the Conditional $MI$ over traditional $MI$.

Table \ref{tab:cmim2_frgc_hog} shows the D-EER for HOG feature with the best method CMIM2. Surprisingly, the shape feature (HOG) reached the best results with the lower D-EER using CMIM-2 and FRGCv2 database. FaceMorpher reached the lower D-EER with 1.8\%  with a BPCER10 of 0.3\%  and BPCER20 of 1.0\%. Conversely, FaceFusion reached the higher D-EER of 5.8\%. The second column shows, the comparison (D-EER) between the HOG results from baseline using all the HOG features versus the proposed method with feature selected from HOG. 

\begin{table}[]
\scriptsize
\centering
\caption{D-EER in \% for the best results reached by CMIM-2 using HOG.}
\label{tab:cmim2_frgc_hog}
\begin{tabular}{|c|c|c|c|}
\hline
FRGCv2 - HOG.  & HOG  / Fea+HOG    & BPCER10 & BPCER20 \\ 
               & (D-EER)           &         &         \\ \hline
FaceFusion     & 27.7/\textbf{5.8} & 3.7     & 7.7   \\ \hline
FaceMorpher    & 29.1/\textbf{1.8} & 0.3     & 1.0   \\ \hline
OpenCV-Morpher & 28.6/\textbf{2.0} & 0.0     & 0.0     \\ \hline
UBO-Morpher    & 26.4/\textbf{4.0} & 2.0     & 4.4   \\ \hline
\end{tabular}
\end{table}

Table \ref{tab:cmim2_feret_lbp} shows the D-EER for uLBP\_ALL feature with the best method CMIM2. For FERET database the best results with the lower D-EER using CMIM-2. FaceMorpher again reached the lower D-EER with 1.3\% with a BPCER10 of 0.3\% and BPCER20 of 1.0\% . Conversely, UBO-Morpher reached the higher D-EER of 9.4\% with a BPCER10 of 2.9\%  and BPCER20 of 13.8\%. The second column shows, the comparison (D-EER) between the uLBP\_all results from baseline using only the fusion of uLBP features versus the proposed method with feature selected from uLBP.

\begin{table}[]
\scriptsize
\centering
\caption{D-EER in \% for the best results reached by CMIM-2 using All\_LBP.}
\label{tab:cmim2_feret_lbp}
\begin{tabular}{|c|c|c|c|}
\hline
FRGCv2-uLBP    & uLBP  / Fea+uLBP & BPCER10 & BPCER20 \\ 
               & (D-EER)          &         &         \\ \hline
FaceFusion     & 14.2/\textbf{9.2} & 7.4   & 20.0   \\ \hline
FaceMorpher    & 17.3/\textbf{1.3} & 0.3   & 1.0   \\ \hline
OpenCV-Morpher & 16.2/\textbf{4.0} & 1.3   & 4.4   \\ \hline
UBO-Morpher    & 11.7/\textbf{9.4} & 2.9   & 13.8   \\ \hline
\end{tabular}
\end{table}

\begin{figure}[]
\centering
\caption{\label{DET2} DET curves for FRGCv2 and FERET using Feature selection method. Top: RAW. Middle: HOG. Bottom: uLBP Fusion.}

\includegraphics[scale=0.14]{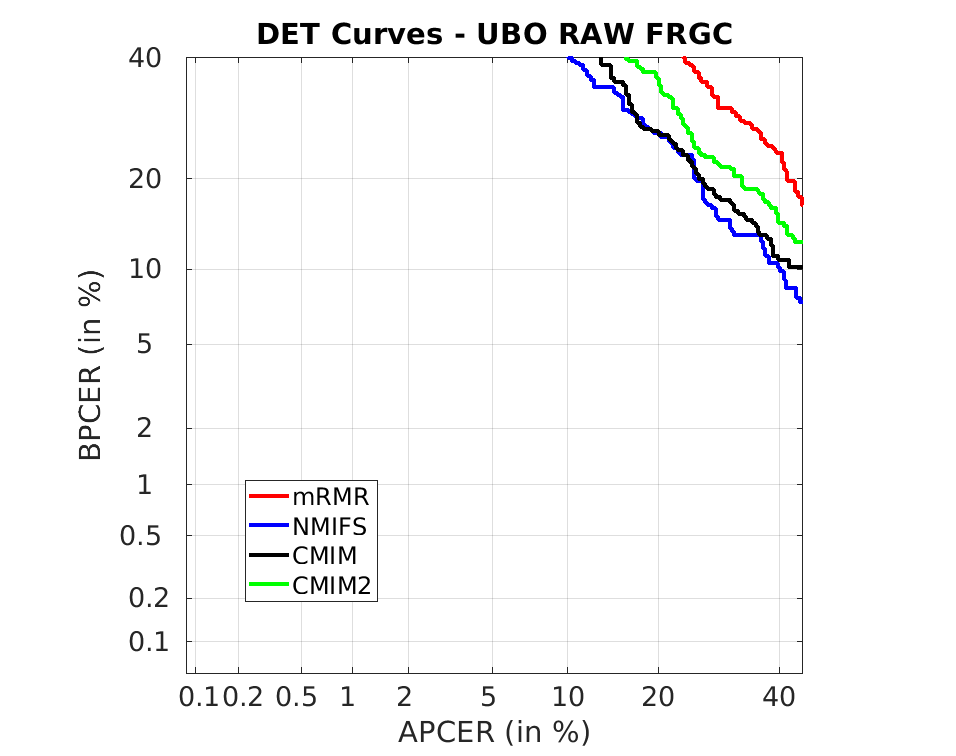}
\includegraphics[scale=0.13]{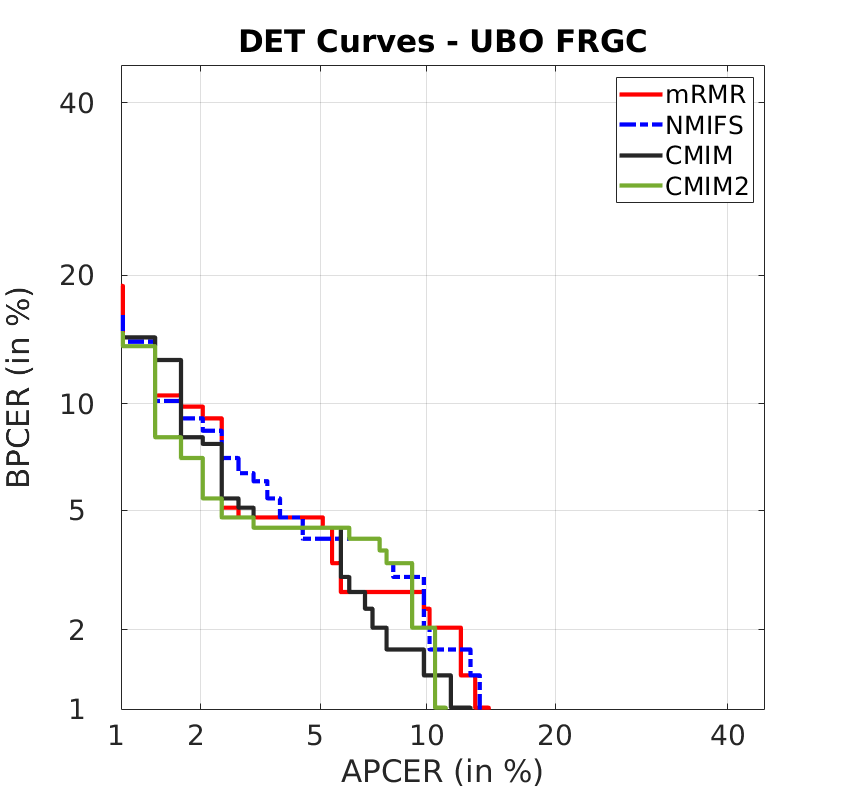}\\
\includegraphics[scale=0.20]{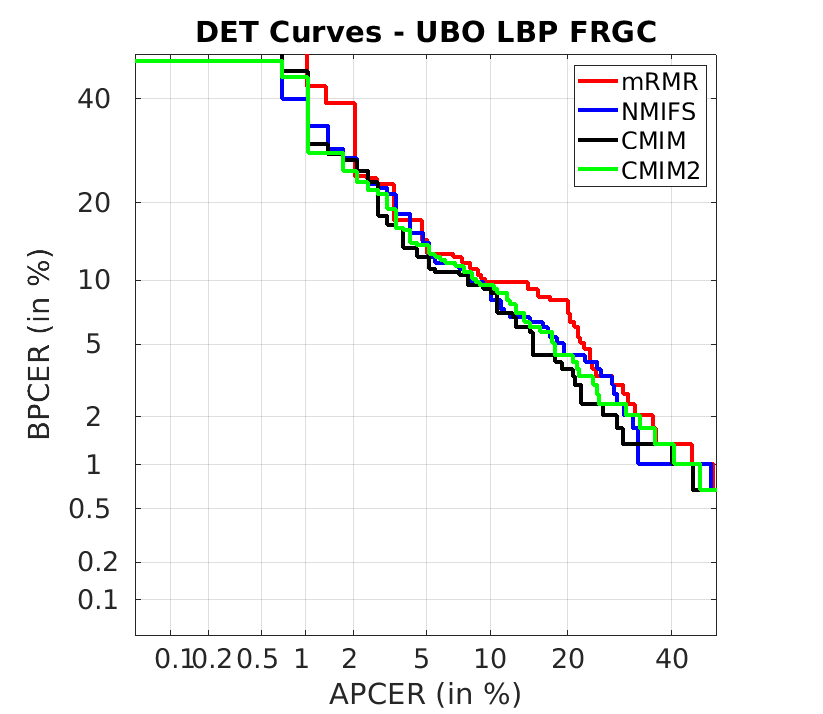}
\includegraphics[scale=0.19]{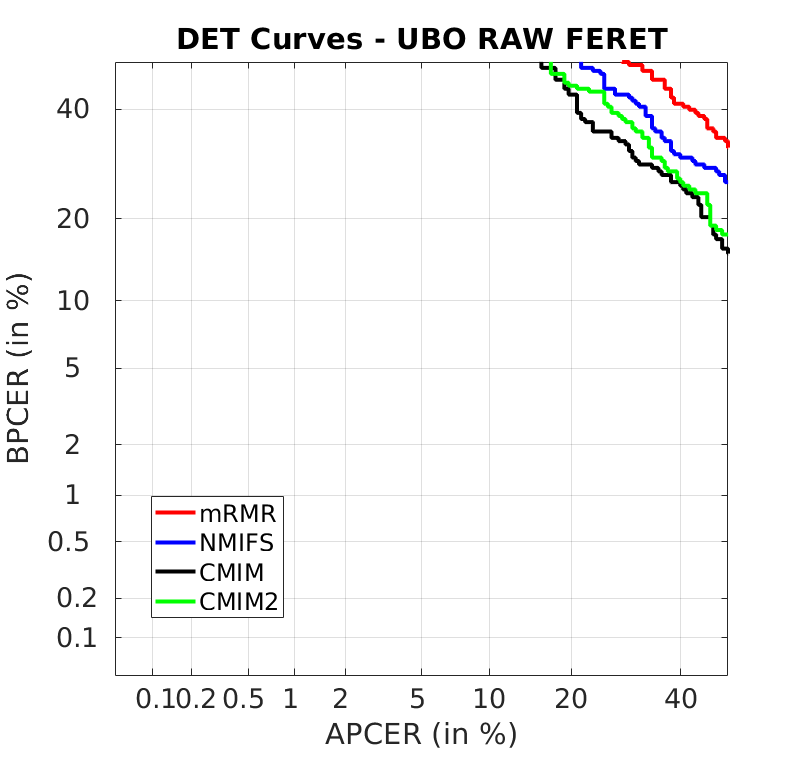}\\
\includegraphics[scale=0.18]{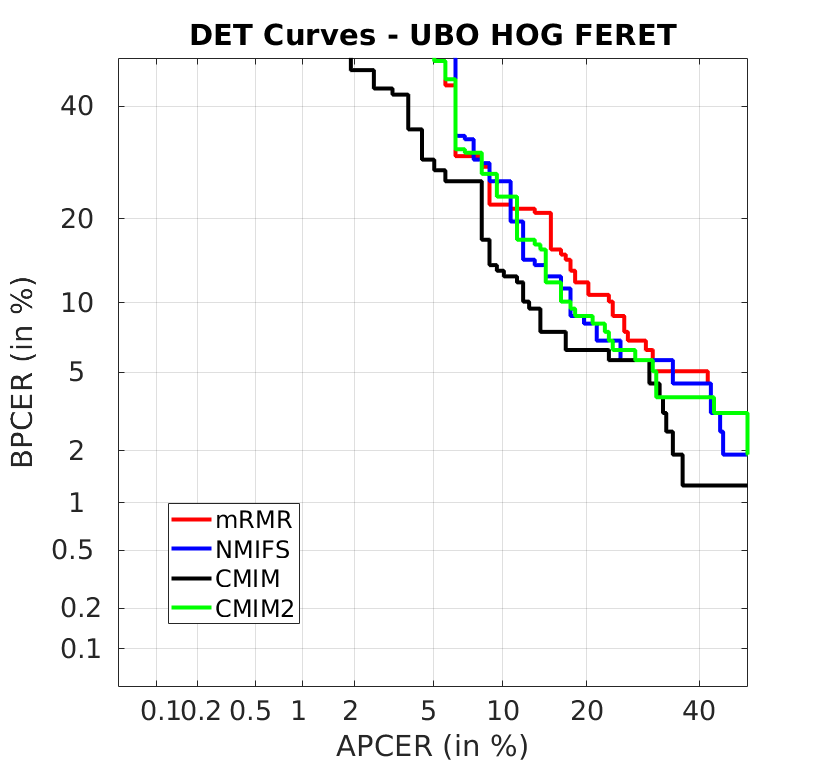}
\includegraphics[scale=0.19]{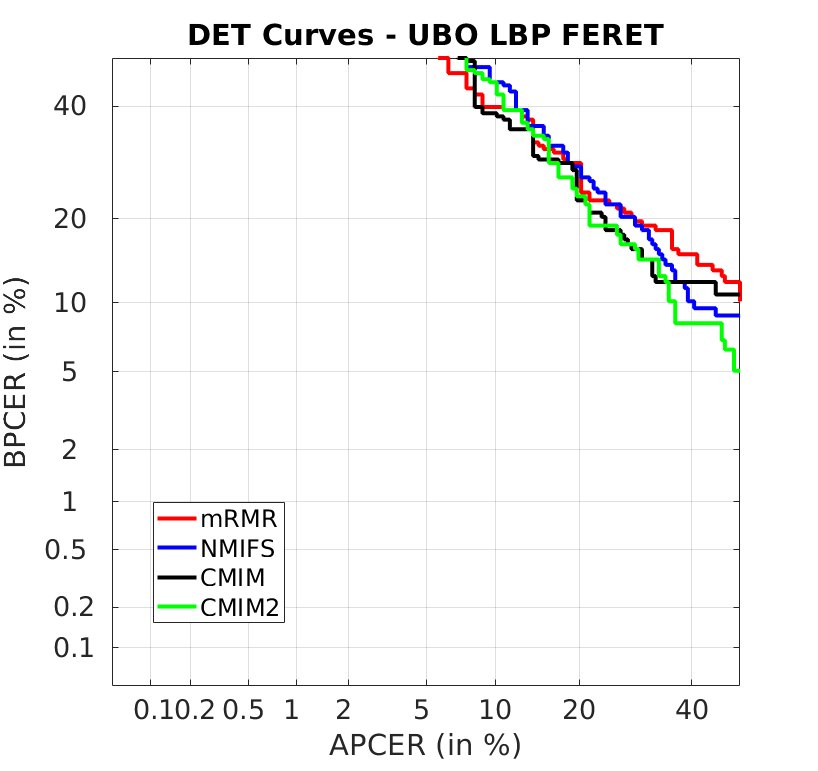}
\end{figure}

Figure \ref{DET2} show the DET curves obtained for the four feature selected method for the three feature selected (Intensity, Texture and Shape). The UBO-Morpher constitutes a high-quality morphing implementation and is applied on FERET and FRGCv2 databases. Conversely, FaceMorpher is the more straightforward method to be detected based on the artefacts present in the images. The features selection applied to intensities values reached the lower results. Even these results improve the baseline, the D-EER are not competitive with the literature. Conversely, uLBP and HOG improve a lot in comparison with the baseline and reached results competitive with the literature as is shown in Tables \ref{tab:cmim2_frgc_hog} and \ref{tab:cmim2_feret_lbp}.

\begin{figure}[]
\centering
\caption{D-EER for comparison of the features selected using CMIM from intensity, shape and texture for FERET database. R: represents RAW. H: represents HOG and L, represents uLBP.}
\includegraphics[scale=0.33]{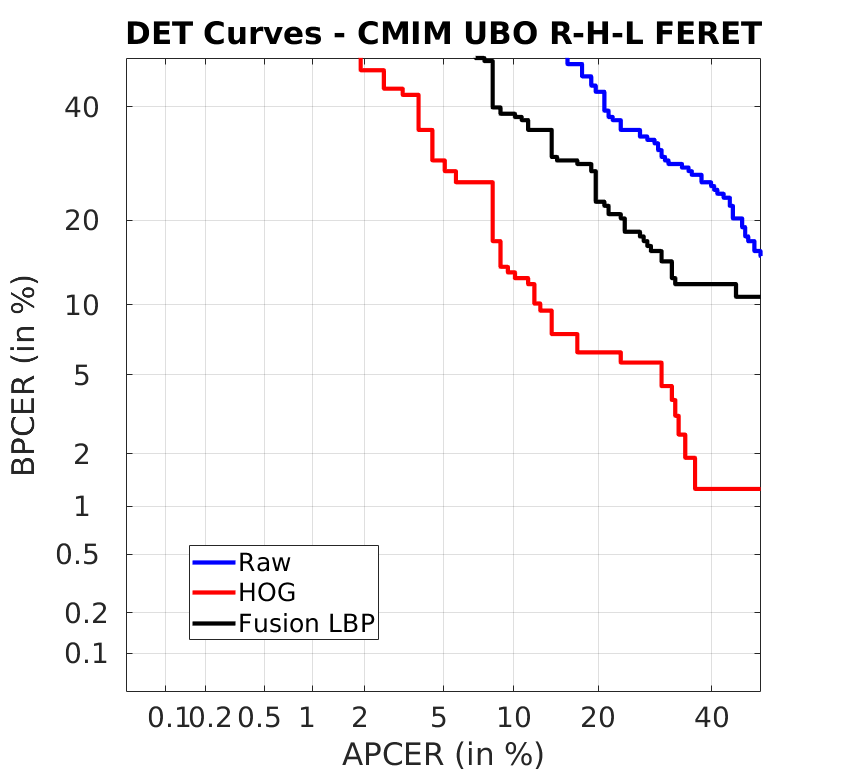}
\label{tab:cmim2_ulbp_RHL_feret}
\end{figure}

\begin{figure}[]
\centering
\caption{D-EER for comparison of the features selected using CMIM from intensity, shape and texture for FRGCv2 database. R: represents RAW. H: represents HOG and L, represents uLBP.}
\includegraphics[scale=0.30]{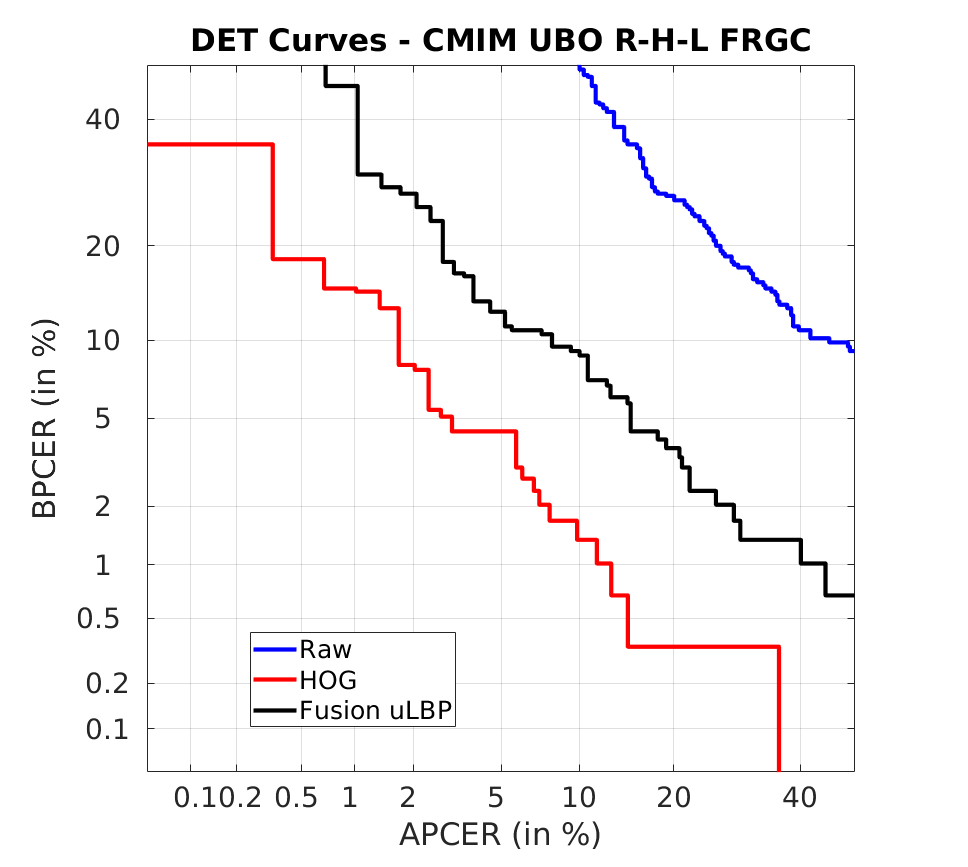}
\label{tab:cmim2_ulbp_RHL_frgc}
\end{figure}

In order to compare and analysed which extracted feature delivers more useful information for the detection task, the Figures \ref{tab:cmim2_ulbp_RHL_feret} and \ref{tab:cmim2_ulbp_RHL_frgc} shows a comparison of FERET and FRGCv2 for best results obtained by CMIM-2 from intensity, shape (HOG) and texture (uLBP). Both figures have shown that HOG reached a lower D-EER in both databases. This result shows that the shape algorithms also can detect morphing images as a complement of textures. The exploration parameters to find the most representative inverse HOG features and their visualisation allows us to improve the results. This is shown in Figure \ref{fig:ihog}.

\section{Visualisation}

\begin{figure}[]
\centering
\textbf{mRMR}\par\medskip 
\includegraphics[scale=0.31]{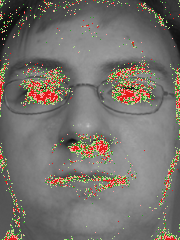}
\includegraphics[scale=0.31]{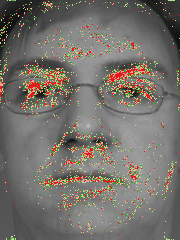}
\includegraphics[scale=0.31]{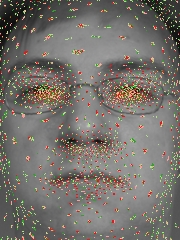}
\includegraphics[scale=0.31]{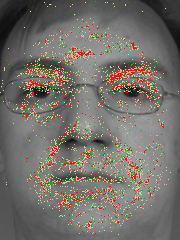}
\\
\textbf{NMIFS}\par\medskip
\includegraphics[scale=0.31]{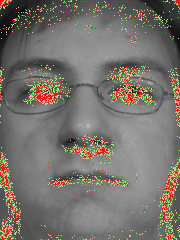}
\includegraphics[scale=0.31]{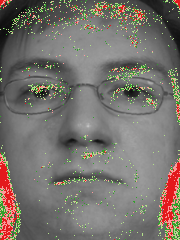}
\includegraphics[scale=0.31]{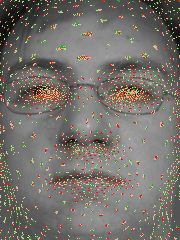}
\includegraphics[scale=0.31]{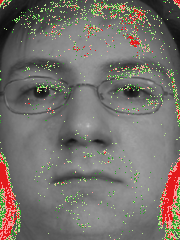}
\\
\textbf{CMIM}\par\medskip
\includegraphics[scale=0.31]{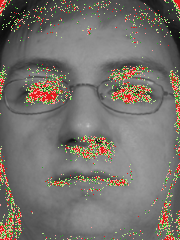}
\includegraphics[scale=0.31]{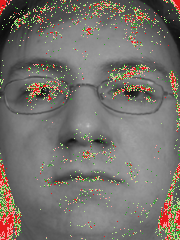}
\includegraphics[scale=0.31]{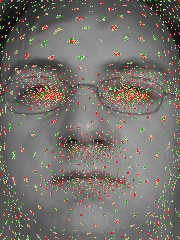}
\includegraphics[scale=0.31]{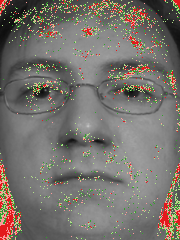}
\\
\textbf{CMIM-2}\par\medskip
\includegraphics[scale=0.31]{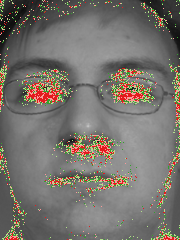}
\includegraphics[scale=0.31]{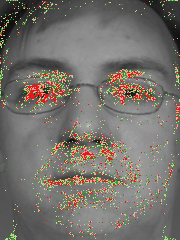}
\includegraphics[scale=0.31]{images/vis2/int_nmifs_ubo.png}
\includegraphics[scale=0.31]{images/vis2/int_nmifs_ubo.png}\\
\textbf{FaceFusion~~~~~~FaceMorph.~~~FaceOpenCV~UBO-Morph.}\par\medskip
\caption{\label{vis_frgc} Localisation of the feature selected by mRMM, NMIFS, CMIM and CMIM2 for different morphing algorithm. Each image shows the best 5.000 features}
\end{figure}

Once we select the best features, it is possible to recover the coordinates of the features into the images. Then, we can visualise the attributes for each method.
Figure \ref{vis_frgc} shows the localisation of the most relevant features for an FRGCv2 random image.
The 5,000 features were divided into five equal parts and assigned to five different colours. The most relevant features from 1 to 1,000 are represented as red pixels. From 1,001 to 2,000 are pink. 2,001 to 3,000 are green. 3,001 to 4,000 are light green, and 4,001 to 5,000 are represented as blue.
It is essential to highlight that the pixels in colours represent the best features selected, which means the most relevant less redundant from the four methods: mRMR, NMIFS, CMIM, and CMIM2, from 1,000 up to 5,000. The CMIM features are distributed in all the images and only concentrate in some areas. The CMIM-2 focalised the features in the most relevant areas. The eyes and the nose areas are selected as relevant to detect morphed images.

\section{Conclusion}
\label{conclusion}

After analysing all the results, we can conclude that morphing based on the FERET database is more challenging to detect than the FRGCv2 database. The leave-one-out protocol is essential to estimate the actual performance of the proposed method. In the literature, the test set typically contains images from the same morphing tools.
The feature selection reduces the number of features used drastically to separate bona fide for morphed images and reduce the D-EER in all the cases. For the feature selected from HOG, the D-EER decreased from 26.4\% (baseline) to 4.0\% for UBO-Morpher, reached a BPCER10 of 2.0\%. For the chosen feature from the fusing uLBP, the D-EER decrease from 11.7\% (baseline) to 8.4\% obtained a BPCER10 of 2.9\%. These results are very competitive with the state of art.
The localisation of the features enabled us to select the most relevant and less redundant features. The nose and eyes are identified as relevant areas in the face for manual analysis of morphed images. This tool may help the border police detect morphing images and address the areas to be analysed for the artefacts. 
In summary, the shape feature (HOG) results outperform the texture performance as is shown in Figures \ref{tab:cmim2_ulbp_RHL_feret} and \ref{tab:cmim2_ulbp_RHL_frgc}. 
In future work, we will apply this method to embedding features extracting from the face-recognition system in order to choose the best features.

\section*{Acknowledgment}
This work is supported by the European Union’s Horizon 2020 research and innovation program under grant agreement No 883356 and the German Federal Ministry of Education and Research and the Hessen State Ministry for Higher Education, Research and the Arts within their joint support of the National Research Center for Applied Cybersecurity ATHENE. 

\section*{Disclaimer}
This text reflects only the author’s views, and the Commission is not liable for any use that may be made of the information contained therein.

\bibliographystyle{IEEEtran}
\bibliography{samples.bib}

\begin{IEEEbiography}[{\includegraphics[width=1in,height=1.25in,clip,keepaspectratio]{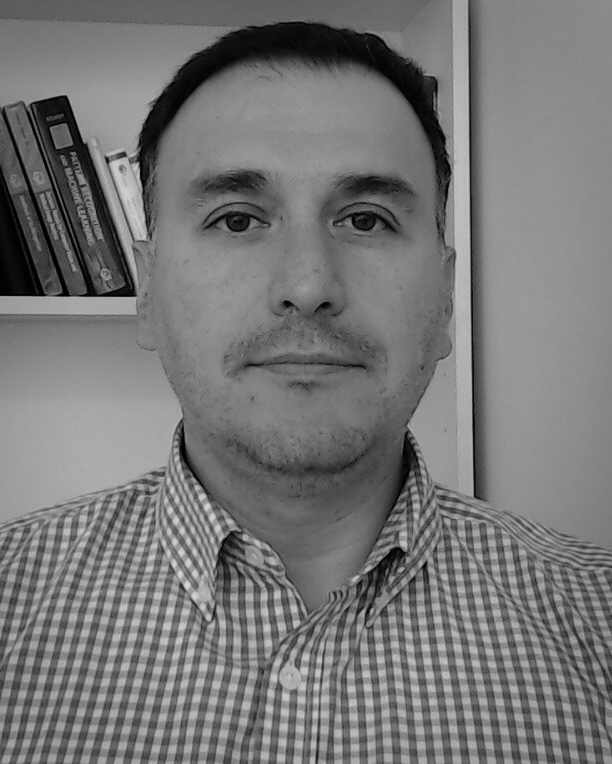}}]{Juan Tapia} received a P.E. degree in Electronics Engineering from Universidad Mayor in 2004, a M.S. in Electrical Engineering from Universidad de Chile in 2012, and a Ph.D. from the Department of Electrical Engineering, Universidad de Chile in 2016. In addition, he spent one year of internship at University of Notre Dame (USA). In 2016, he received the award for best Ph.D. thesis. From 2016 to 2017, he was an Assistant Professor at Universidad Andres Bello. From 2018 to 2020, he was the R\&D Director for the area of Electricity and Electronics at Universidad Tecnologica de Chile - INACAP. He is currently a Senior Researcher at Hochschule Darmstadt(HDA), and R\&D Director of TOC Biometrics. His main research interests include pattern recognition and deep learning applied to iris/face biometrics, vulnerability analysis, morphing, feature fusion, and feature selection.
\end{IEEEbiography}

\begin{IEEEbiography}[{\includegraphics[width=1in,height=1.25in,clip,keepaspectratio]{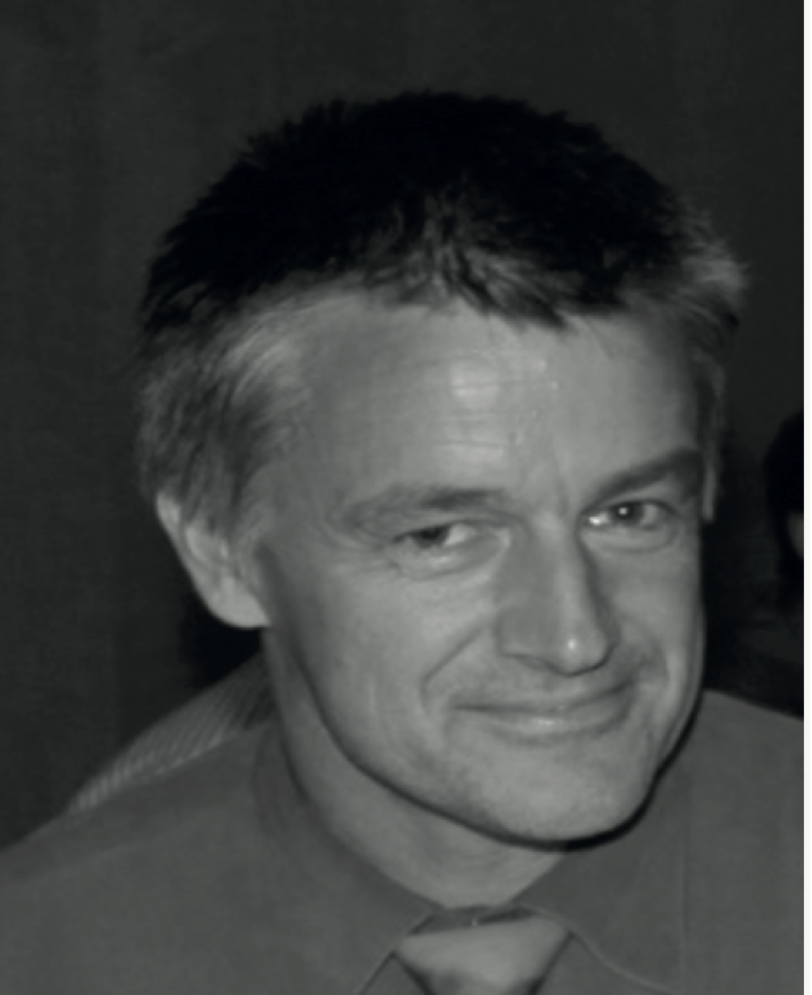}}]{Christoph Busch} is member of the Department of Information Security and Communication Technology (IIK) at the Norwegian University of Science and Technology (NTNU), Norway. He holds a joint appointment with the computer science faculty at Hochschule Darmstadt (HDA), Germany. Further he lectures the course Biometric Systems at Denmark’s DTU since 2007. On behalf of the German BSI he has been the coordinator for the project series BioIS, BioFace, BioFinger, BioKeyS Pilot-DB, KBEinweg and NFIQ2.0. In the European research program he was initiator of the Integrated Project 3D-Face, FIDELITY and iMARS. Further he was/is partner in the projects TURBINE, BEST Network, ORIGINS, INGRESS, PIDaaS, SOTAMD, RESPECT and TReSPAsS. He is also principal investigator in the German National Research Center for Applied Cybersecurity (ATHENE). Moreover Christoph Busch is co-founder and member of board of the European Association for Biometrics (www.eab.org) that was established in 2011 and assembles in the meantime more than 200 institutional members. Christoph co-authored more than 500 technical papers and has been a speaker at international conferences. He is member of the editorial board of the IET journal.
\end{IEEEbiography}


\end{document}